\documentclass{article}
\pdfoutput=1

\usepackage{microtype}
\usepackage{graphicx}
\usepackage{subfigure}
\usepackage{booktabs} 

\usepackage[accepted]{icml2024}

\usepackage{amsmath}
\usepackage{amssymb}
\usepackage{mathtools}
\usepackage{amsthm}

\usepackage[capitalize,noabbrev]{cleveref}

\theoremstyle{plain}

\theoremstyle{definition}

\theoremstyle{remark}

\usepackage[textsize=tiny]{todonotes}

\usepackage[algo2e,ruled,vlined]{algorithm2e}
\usepackage{algorithmic}
\usepackage{algorithm}
\usepackage{eqparbox}

\usepackage{mathtools}
\usepackage{verbatim}
\usepackage{multirow}
\usepackage{multicol}
\usepackage{wrapfig}
\usepackage{amssymb}

\icmltitlerunning{Learning Object Permanence from Videos}

\begin{document}

\twocolumn[
\icmltitle{Learning Object Permanence from Videos via Latent Imaginations} 



\icmlsetsymbol{equal}{*}

\begin{icmlauthorlist}
\icmlauthor{Manuel Traub}{equal,yyy}
\icmlauthor{Frederic Becker}{equal,yyy}
\icmlauthor{Sebastian Otte}{yyy}
\icmlauthor{Martin Butz}{yyy}
\end{icmlauthorlist}

\icmlaffiliation{yyy}{Neuro-Cognitive Modeling Group, University of Tübingen, Germany}

\icmlcorrespondingauthor{Manuel Traub}{manuel.traub@uni-tuebingen.de}
\icmlcorrespondingauthor{Frederic Becker}{frederic.becker@uni-tuebingen.de}

\icmlkeywords{compositional representation learning, neuro-cognitive modeling, development psychology, machine Learning}

\vskip 0.3in
]



\printAffiliationsAndNotice{\icmlEqualContribution} 

\def\locil*{Loci-Looped}
\def\loci*{Loci-v1}
\newcommand{\rebuttal}[1]{\textcolor{red}{#1}}

\begin{abstract}
While human infants exhibit knowledge about object permanence from two months of age onwards, deep-learning approaches still largely fail to recognize objects' continued existence. 
We introduce a slot-based autoregressive deep learning system, the looped location and identity tracking model \locil*, which learns to adaptively fuse latent imaginations with pixel-space observations into consistent latent object-specific what and where encodings over time.  
The novel loop empowers \locil* to learn the physical concepts of object permanence, directional inertia, and object solidity through observation alone. 
As a result, \locil* tracks objects through occlusions, anticipates their reappearance, and shows signs of surprise and internal revisions when observing implausible object behavior.
Notably, \locil* outperforms state-of-the-art baseline models in handling object occlusions and temporary sensory interruptions while exhibiting more compositional, interpretable internal activity patterns.
Our work thus introduces the first self-supervised interpretable learning model that learns about object permanence directly from video data without supervision.

\end{abstract}

\section{Introduction}
State-of-the-art Artificial Intelligence (AI) systems achieve impressive performance in object detection, instance segmentation, and object tracking tasks \citep{he_mask_2017, wang_yolov7_2022}. 
Yet these systems hardly develop any intuitive physical knowledge, such as object permanence (i.e., objects continue to exist when hidden) or directional inertia (i.e., objects continue their motion unless acted on by an external force) \citep{Weihs:2022}.
This understanding, however, is key to interact with our environment flexibly and effectively in a goal-directed manner \citep{butz_towards_2021,lake_building_2016,spelke_core_2007,spelke_origins_1992}.

During infancy, humans learn physical concepts in the form of expectations about how objects behave, forming object files and performing physical reasoning \citep{aguiar_25-month-old_1996,Lin:2022,summerfield_expectation_2009}. These expectations have been explicitly probed with the Violation-of-Expectation (VoE) paradigm \citep{baillargeon_object_1985}: infants are shown videos that either adhere to (e.g., an occluded object reappears) or violate (e.g., an occluded object vanishes) a physical concept (e.g., object permanence and inertia), while monitoring their gaze behavior. 
Once the physical concept has developed, infants look longer at the physics violating events compared to similar plausible ones.
It appears, that infants learn internal event-predictive models. These model encode the entities and interactions that are unfolding within and between events.
Further during development, we then use these models to infer currently unfolding events and to anticipate the consequences, which enables us to intearct with our world in an anticipatory, pro-active manner \citep{clark_whatever_2013,butz_event-predictive_2021, Butz:2008h, Butz:2017book,den_ouden_how_2012,Lin:2022,Zacks:2007,Baldwin:2021,Kuperberg:2021}.

The challenge to model the development of object permanence with artificial neural network reaches back to experiments in the last century with recurrent neural networks \citep{Munakata:1997}.
It is closely related to the binding problem, which asks for the ability to segregate streams of information, represent entities independently, and enabling their flexible composition into interaction structures \citep{Butz:2017book,Greff:2020}.
In a recent study working with actual video data, \citet{piloto_intuitive_2022} leveraged the idea of predictive coding. 
They first trained a deep learning model on next-frame prediction tasks and then assessed the model's understanding of intuitive physics using the VoE paradigm.
Their results indicate that their model had learned multiple physical concepts.
However, although the model was trained in a self-supervised manner, it received supervised information regarding the location and identity of each object in the scene in the form of object-respective ground truth masks. 
As a result, the challenge of learning object permanence was side-stepped via the provided object masks.
Thus, 
solutions to the segmentation and tracking problems, that is, the segregation problem in the terminology of \citet{Greff:2020}, were provided a priori.



In contrast, compositional scene representation models \citep{traub_learning_2023, kipf_conditional_2022, elsayed_savi_2022} do learn to both segment a scene into individual objects and track them over time in a self-supervised manner. Most of these models however do 
not take (partially) hidden objects into account when representing a scene or do not provide any option to decode and assess representations of hidden objects (see Section~\ref{sec:relatedwork}). An exception is the recently introduced location and identity tracking model \citep{traub_learning_2023}, named Loci (here referred to as \loci*), which showed superior performance in the CATER benchmark \citep{girdhar_cater_2020}, where objects need to be tracked that are contained in another object. 
Thus, \loci* can learn the concept of containment but partially struggles with more general scenarios that are probing object permanence and inertia knowledge.  
Accordingly, here we test three state-of-the-art compositional scene representation models (SAVi \citep{kipf_conditional_2022}, \loci* \citep{traub_learning_2023} and G-SWM \cite{lin_gswm_2020}) in a benchmark scenario. 
We find that all models struggle even with rather simple occlusion scenarios. 
These results show that a model that would learn object permanence end-to-end from videos is still missing.

To overcome this limitation, we propose \locil*, which augments the object-centric encoder-transition-decoder architecture of \loci* with the ability to fuse latent temporal imaginations with pixel-space observations into consistent, compositional scene percepts. While an outer sensory loop allows \locil* to build and update representations of visible objects, the novel inner loop allows to imagine object-centric latent state dynamics---much like the dreamer architecture \citep{hafner_dream_2020,Wu:2023daydreamer}. 
The inner loop thus enables \locil* to simulate the state of temporarily hidden objects over time.
Importantly, \locil* learns without supervision to adaptively fuse external, sensory information with internal, anticipated information for each object individually via a parameterized percept gate. 
Our experiments show that \locil* learns to imagine the trajectory of temporarily hidden objects thereby developing the principles of object permanence, directional inertia, and object solidity. Our ablation studies confirm that the inner-loop and the flexible control of it are key for learning this behavior.

The main contribution of our work is the development of a fully unsupervised, autoregressive learning system, called \locil*, that learns to 
\begin{itemize}
    \setlength\itemsep{-0.1em}
    \item adaptively and selectively fuse internal beliefs with external evidence;
    \item track moving objects over time, particularly also when they are hidden over extended periods of time or when blackouts temporarily conceal visual information; 
    \item show surprise when objects do not reappear where and when they should;
    \item form concepts of object permanence, directional inertia, and object solidity from scratch---an ability that has not yet been achieved by any other fully self-supervised learning system;
    \item generate interpretable latent activity patterns that reveal how \locil* infers compositional, slot-based, object-specific encodings and how it adaptively fuses internal anticipated with sensory-based object-respective information over time.
\end{itemize}


\section{Related Work}
\label{sec:relatedwork}
Recently various approaches in the field of compositional scene representation learning have been proposed. All methods decompose a scene into multiple components and represent the scene by a composition of the components. Ideally this decomposition corresponds to semantically meaningful image parts (typically objects). Following \citet{yuan_compositional_2023}, we give a brief overview of the main characteristics of five recent models, namely Slot Attention \citep{ding_attention_2021}, SAVi \citep{kipf_conditional_2022}, G-SWM \citep{lin_gswm_2020}, MONet \citep{burgess_monet_2019}, and \loci* \citep{traub_learning_2023}.

\textbf{Layer Composition}
In the decoding process, all models aim to reconstruct the current scene from a stack of individual layers, each focusing on individual objects. To merge these layers into one reconstruction, two approaches are commonly used. In the first approach, exemplified by MONet, the value of a pixel is only determined by one layer that is sampled based on spatial mixture weights. Alternatively, SlotAttention, SAVi, G-SWM, and \loci* reconstruct the scene by summing over all layers while weighting the contribution of each layer in each pixel individually.

\textbf{Scene Decoding}
Methods also differ in the decoding process.
SlotAttention and SAVi apply layer summation directly in the decoding process, producing a single reconstruction that depicts only visible object parts. Hidden object parts can be part of their latent object representation but are not decoded. Alternatively, G-SWM mimics a more holistic scene decoding. The decoding process produces one reconstruction per object showing complete object shapes, which are then ordered based on depth variables. 
In this work, we reconstruct both full object shapes and only the visible parts, thus combining both approaches. 

\textbf{Object Representation}
Objects are typically encoded as low-dimensional vectors, serving as an information bottleneck that encourages latent object compression. 
Approaches such as SAVi and SlotAttention sample these encodings from a prior distribution as generative models. In contrast, G-SWM, MONet and \loci* do not depend on a prior distribution.

\textbf{Object Counting}
Methods vary in their capacity to explicitly count and represent the number of objects in a scene. Unlike the other approaches, G-SWM and \loci* can flexibly adjust the number of components that are used to represent the scene. Consequently, they can explicitly capture and represent the actual number of objects present.

\textbf{Attention Mechanism}
The simulation of dynamic interactions between scene components is commonly achieved through the use of attention mechanisms. Attention can be employed to model relations between rectangular image regions, such as object bounding boxes, or to capture relationships between arbitrary-shaped image regions based on the latent, self-organizing object representations. The latter approach is used by SlotAttention, SAVi, MONet, and \loci*. 

\textbf{Intuitive Physics}
Recently, the PLATO \citep{piloto_intuitive_2022} and the ADEPT  \citep{smith_modeling_2019} models have gained attention for learning the physical concepts of object permanence, solidity, and continuity.  However, while both models adopt object-centric architectures, they rely on pre-existing segmentation information and supervision. A comprehensive review of these models can be found in Appendix~\ref{app:related}.

\section{Method}
We give a brief introduction to \loci* \citep{traub_learning_2023} baseline architecture including its formalization. We then introduce our novel developments defining \locil*. 
Appendix~\ref{app:method} provides further details. 

\subsection{\loci*}
\loci* consists of three main components: an encoder module that parses visual information into object representations, a transition module that projects these representations into the future, and a decoder module that reconstructs a visual scene from this prediction. Each of the three components comprises $k$ slots that share their weights. Each slot is dedicated to process one object. It may stay empty when more slots than objects are available.

The ResNet-based, slotted encoder module receives the current frame $I^t$, the previous prediction error $E^t$, a background mask $\hat{M}^t_{bg}$ as well as slot-specific predictions of position $\hat{Q}^t_k$, visibility mask $\hat{M}^{t,v}_k$, RGB object reconstruction $\hat{R}^t_k$, and the summed visibility mask of the remaining slots $\hat{M}^{t,s}_k$.
Positions are encoded as isotropic Gaussians in pixel space, masks as grayscale images.
The encoder produces Gestalt codes $\tilde{G}^t_k$ and positional codes $\tilde{P}^t_k$ as output. Gestalt codes encode shape and surface patterns, while positional codes include object location ($x_k,y_k$), size ($\sigma_k$), and priority ($\rho_k$).

The transition module predicts the encodings at the next timestep, namely $\hat{G}^{t+1}_k$ and $\hat{P}^{t+1}_k$. It implements a combination of slot-wise recurrent layers to model object dynamics and across-slot attention layers to model object interactions.
In contrast to the PLATO model \citep{piloto_intuitive_2022}, the recurrent layers do not receive a history of object states depicting previous object dynamics. 
Following the transition module, the Gestalt codes are binarized, creating an information bottleneck that biases the slots to develop factorized compositional encodings of entities. 

The decoder module then reconstructs the predicted scene from $\hat{G}^{t+1}_k$ and $\hat{P}^{t+1}_k$ and a provided image of the scene background, which marks the only supervised model input. 
For each slot, a ResNet architecture produces the predicted RGB object reconstruction $\hat{R}^{t+1}_k$, visibility mask $\hat{M}^{t+1,v}_k$, and position $\hat{Q}^t_k$. All slot outputs are unified in the prediction $\hat{R}^{t+1}$ by taking the sum over the RGB object reconstructions weighted by the visibility masks and the background mask. Along with the next input frame $I^{t+1}$ the prediction serves to generate prediction error $E^{t+1}$. 
This process repeats in each timestep. 

\begin{figure}[t]
    \centering
    \includegraphics[width=0.99\linewidth]{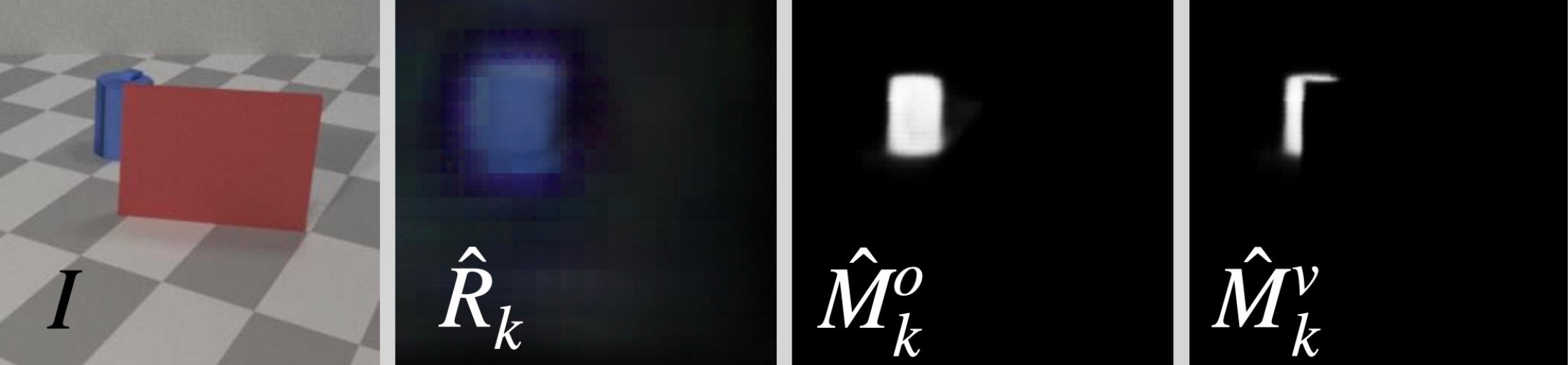}
    \caption{ The object and visibility mask enable an interpretable holistic scene understanding in Loci-Looped. \textit{From left to right}: Current video frame, reconstructed RGB object, object mask and visibility mask of slot $k$ depicting the blue object.}
    \label{fig:inputs}
\end{figure}
\hspace{-4cm}

\subsection{\locil*}
\subsubsection{Object Mask \footnote{Utilizing the identical decoding procedure, object masks and occlusion states could be generated for a variety of compositional scene representation models, including SAVi and Slot Attention.}}
\label{sec:objectmask}
Visibility masks outputted by most compositional scene representation models exclusively depict visible components of objects. To enable a holistic scene understanding, we introduce an extra mask that is designed to encode entire object shapes, which serves as an additional input to the encoder of \locil*. To compute this mask, we assume that only slot-object $k$ is in the scene, ignoring the remaining slots. Consequently, in the decoding process slot $k$ only competes with the background for visibility yielding object mask 
\begin{equation}
    \label{eq:objectmask} 
    M^{t,o}_k = \frac{\mathrm{exp}(M^{t}_k)}{\mathrm{exp}(M^{t}_k) + \mathrm{exp}(M^{t}_{bg})},
\end{equation}
where $M$ is generated by the decoder. 

\begin{figure*}[h]
    \centering
    \includegraphics[width=0.9\linewidth]{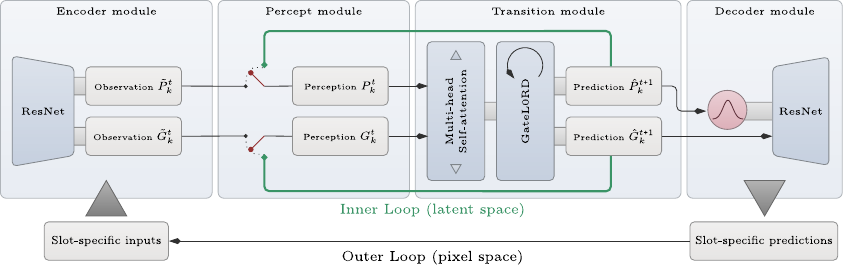}
    \caption{The slot-wise processing architecture of \locil*. Predictions are made available on two routes. First, through an outer loop in pixel-space, which enables continuous visual object tracking over time; second, through an inner loop, which enables the generation of latent temporal imaginations.
    }
    \label{fig:Loci}
\end{figure*}

\subsubsection{Occlusion State}
The introduction of the object mask enables \locil* to determine the degree of occlusion for each object. We calculate the occlusion state $O^t_k$ as follows:
\begin{equation}
\label{eq:occlusionstate}
 O^t_k = 1- \frac{\sum_{i,j} [M^{t,v}_k(i,j) > \theta]}{\sum_{i,j} [M^{t,o}_k(i,j) > \theta] + c}, 
\end{equation}
where $\theta$ is a threshold value, which we set to $0.8$, and $c$ is a small constant. By counting the number of pixels larger than threshold $\theta$, the denominator determines the total area of the object, while the numerator determines the visible area of the object. The occlusion state ranges from 0 (fully visible) to 1 (fully occluded), allowing \locil* to explicitly represent the state of occlusion, increasing interpretability and serving as input to the percept gate controller.

\subsubsection{Percept Gate}
\loci*'s object tracking approach draws inspiration from Kalman filtering, which iteratively predicts object state changes and then adaptively fuses these predictions with current observations \citep{kalman_new_1960}. Accordingly, \loci* predicts the next object states, decodes them into pixel space and then uses these predictions along with the current frame to produce new object states (see Figure~\ref{fig:Loci}; outer loop). While the Kalman filter separates the steps of observation and information fusion, \loci* observes and fuses jointly and implicitly during the encoding process. This is advantageous when fusing pixel-based information (e.g., combining hidden and visible object parts). However, when the model needs to fully maintain its own predictions because the current frame does not provide new information (e.g., during full occlusion), the encoding process via the outer loop becomes disruptive.
As an alternative, recent work from model-based reinforcement learning advocates the efficiency and precision of predicting directly in latent space \citep{hafner_learning_2019,hafner_dream_2020, ha_world_2018}. Latent world models can be used to imagine how a scene will unfold while not being provided with new observations, which is the case during temporary occlusions or blackouts. 
Therefore, 
we introduce an inner processing loop in \locil*, which enables the model to propagate internal imaginations over time in latent space 
(see Figure~\ref{fig:Loci}; inner loop).

\begin{table*}[h]
  \centering
  \caption{Training objectives used by \loci* and \locil*, where $\mathrm{BCE}$ denotes the pixel-wise binary cross-entropy loss, $\mathrm{D}$ denotes the decoder, $p_0$ the image center and $\Theta$ the Heaviside function.}
  \label{table:losses}

  \resizebox{\textwidth}{!}{\begin{minipage}{1.1\textwidth}
  \centering
  
    \begin{tabular}{llccll}
    \toprule

    Name &  Type & \loci* & \locil* & Term & Weighting\\
    \midrule
    Next-Frame Prediction & Loss & \checkmark & \checkmark & $ \mathrm{BCE}(I^{t+1},\mathrm{D}(\hat{G}^{t+1},\hat{P}^{t+1})) $ & 1 \\

    Input-Frame Reconstruction & Loss & - &\checkmark & $ \mathrm{BCE}( I^t, \mathrm{D}(\tilde{G}^t,\tilde{P}^t)) $ & 0.33\\ 

    Gestalt Change & Reg & \checkmark & \checkmark & $ \sum_k \bigl[ \mathrm{D}_k(p_0,G_k^{t}) - \mathrm{D}_k(p_0,\hat{G}_k^{t+1}) \bigr]^2$ & 0.1\\

    Position Change & Reg & \checkmark & \checkmark & $\sum_k \bigl[ P_k^{t} -  \hat{P}_k^{t+1} \bigr]^2$ & 0.01\\

    Percept Gate Openings & Reg  & - & \checkmark & $ \sum_k (\Theta(\alpha^{t,G}_k) + \Theta(\alpha^{t,P}_k)) $ & 5e-06\\ 

    Object Permanence &  Reg  & \checkmark & - & $\sum_k \bigl[ D_k(P_k^{t}, G_k^{t}) -  D_k(P_k^{t}, \overline{G}_k^{t}) \bigr]^2$ & - \\

    \bottomrule
    \end{tabular}

    \end{minipage}}
\end{table*}

Similar to the Kalman filter, we equip the model with the ability to linearly interpolate between the current observations and the last predictions. Formally, the current object states $\smash{G^{t}_k}$, $\smash{P^{t}_k}$ become a linear blending of the observed object states $\smash{\tilde{G}^{t}_k}$, $\smash{\tilde{P}^{t}_k}$ and the predicted object states $\smash{\hat{G}^{t}_k}$, $\smash{\hat{P}^{t}_k}$:
\begin{gather}
  G^{t}_k = \alpha^{t,G}_k \tilde{G}^{t}_k + (1-\alpha^{t,G}_k) \hat{G}^{t}_k  
\\
 P^{t}_k = \alpha^{t,P}_k \tilde{P}^{t}_k + (1-\alpha^{t,P}_k) \hat{P}^{t}_k
\end{gather}
The weighting $\alpha$ is specific for each Gestalt and position code in each slot $k$. Importantly, \locil* learns to regulate the two percept gates on its own in a fully self-supervised manner. It learns an update function $g_{\theta}$, which takes as input the observed state $\tilde{S}^{t}_k$, the predicted state $\hat{S}^{t}_k$, and the last positional encoding $P^{t-1}_k$:
\begin{equation}
    (z_k^{t,G}, z_k^{t,P}) = g_{\theta}(\tilde{S}^{t}_k, \hat{S}^{t}_k, P^{t-1}_k) + \varepsilon,
\end{equation}
where a state comprises the Gestalt encoding, the positional encoding, and the occlusion state. By adding noise $\varepsilon$ sampled from a Gaussian with a fixed standard deviation to the function $g_{\theta}$, 
the gates tend to either close or open, rather than remaining partially open \citep{gumbsch_sparsely_2022}.
We model $g_{\theta}$ with a feed-forward network (see Appendix~\ref{app:updategate}). To be able to fully rely on its own predictions, \locil* needs to be able to fully close the gate by setting $\alpha$ exactly to zero. We therefore use a rectified hyperbolic tangent to compute $\alpha$:
\begin{equation}
    (\alpha^{t,G}_k, \alpha^{t,P}_k) = \mathrm{max}(0, \mathrm{tanh}((z_k^{t,G}, z_k^{t,P}))).
\end{equation}
To encourage robust world models without the reliance on continuous external updates, we impose an $L_0$ loss on gate openings (see Section~\ref{sec:training}) encouraging the sparse use of observations.
The introduction of the percept gate enables \locil* to control its perception flexibly fusing predictions with observations, essentially estimating their relative information values. 

\subsection{Training}
\label{sec:training}
We adopt the training procedure of \loci*. 
\locil* is trained in a fully unsupervised manner and undergoes end-to-end training, utilizing the rectified Adam optimization \citep{liu_variance_2021} in conjunction with truncated backpropagation through time (see Appendix~\ref{app:trainingprocedure} for details). 

\subsection{Loss functions}
\label{sec:losses}
A complete list of the training losses used is presented in Table~\ref{table:losses}. Compared to \loci*, we dispense the use of an object permanence loss, which explicitly facilitated the maintenance of object representations in case of occlusions. Instead, \locil* learns the concept of object permanence autonomously. Furthermore, it is worth noting that the percept gates do not only control the forward information flow, but also the backward flow of gradients. When the percept gates are closed, the error signal is only backpropagated to the transition module but not to the encoder module, which could lead to its degeneration. To avoid this, we incorporate a reconstruction loss in \locil* that is directly derived from the current observations. We provide ablations studies on these loss functions in Appendix~\ref{app:ablations}.

\section{Experiments and Results}
We evaluate \locil* in three experiments demonstrating that it learns (i) to reliably track objects through occlusion, (ii) the concept of object permanence by anticipating the reappearance of occluded objects in VoE-like settings, (iii) to handle situations where visual data is temporarily missing and, (iii) to imagine how scenes unfold.

\paragraph{Baselines} 
We compare \locil* against three state-of-the-art models namely, \loci*, SAVi \citep{kipf_conditional_2022} and G-SWM \cite{lin_gswm_2020}. Additionally, we utilize two ablation models. In the first one, we specifically ablate the percept gate (inner loop) by training a version of \locil* that can only make use of the outer loop, labelling this variant Loci-Unlooped. In the second one, we perform an ablation on the parameterized update function $g_{\theta}$ controlling the percept gate. This is achieved by switching to the inner loop directly proportionally to the perceived occlusion state of each object during testing (i.e. $\alpha^t_k = 1-O^t_k$). Consequently, this model uses the inner loop when objects are occluded and the outer loop when objects are visible, terming this variant Loci-Visibility.

\begin{figure*}[h]
    \centering
    \includegraphics[width=0.9\linewidth]{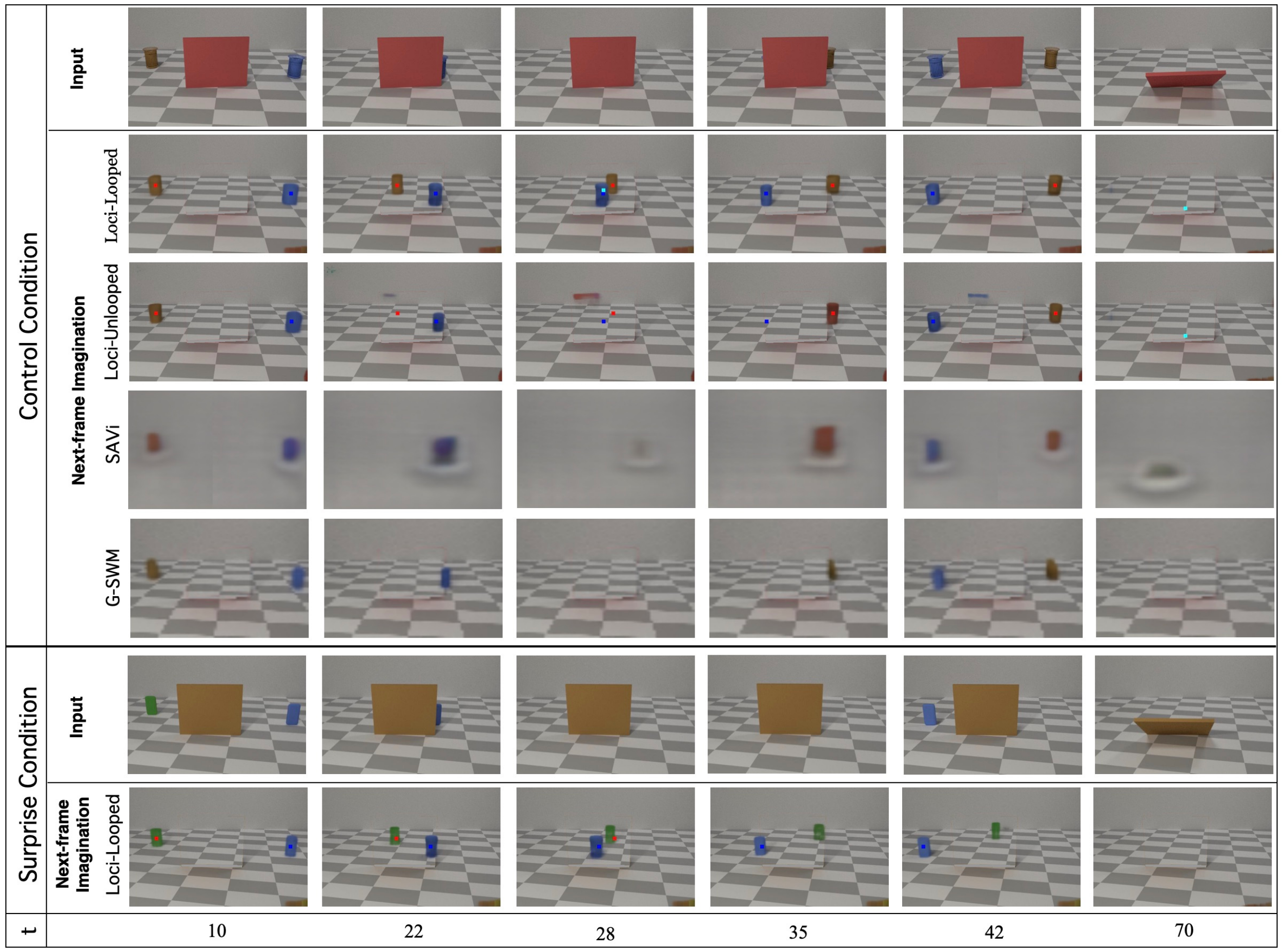}
    \caption{\locil* maintains stable object precepts of the occluded objects. \textit{Control Condition:} Two objects traverse the scene and both objects reappear. \textit{Surprise Condition:} Two objects traverse the scene, the blue object reappears while the green object vanishes. \textit{Next-frame Imagination:} The model's imagination on how the scene unfolds behind the occluder, generated by applying layer summation without the occluder slot. The colored dots show the GT positions of the objects.}
    \label{fig:controledemo}
\end{figure*}

\subsection{Tracking Objects through Occlusion}

\paragraph{Training set}
We train on the ADEPT \citep{smith_modeling_2019} dataset. The training set contains 1000 synthetic videos displaying up to 7 solid objects traversing the scene with constant speed and direction. The training set shows physically plausible dynamics including partial and full object occlusions, while excluding any other object interactions (e.g., collisions).

\paragraph{Test set}
\label{sec:testscenario}
We use 35 videos of the ADEPT vanish scenario as test set. This scenario starts with a large screen placed in the center of the scene. Then one or two objects enter the scene from opposite directions, disappear behind the screen, traverse the area behind the screen while hidden, reappear on the other side of the screen, and finally exit the scene (see Fig~\ref{fig:app:controldemo}). The traversing objects are not visible for 10.3 frames on average which equals 25.0\% of their total time being present.

\paragraph{Metric}
We evaluate the performance of the models with respect to two key abilities. First, we quantify how well the models detect objects and identify them temporally consistently using Multiple Object Tracking Accuracy (MOTA) \citep{bernardin_evaluating_2008}. Second, we quantify the model's tracking error as the distance between estimated object positions and the true object positions. The estimated object positions can be easily extracted, as \locil* represents positional information explicitly. To extract object positions from the SAVi and G-SWM model, we first calculate object masks for each slot (see Section~\ref{sec:objectmask}) and then determine the center of them. Importantly, temporarily occluded objects are included in both metrics (see Appendix~\ref{app:exp1} for details).

\paragraph{Results}
As shown in Figure~\ref{fig:controledemo}, only Loci-Looped maintains stable object representations throughout the occlusion phase and precisely imagines the trajectory of the occluded objects. The average tracking error and the MOTA are listed in Table~\ref{table:tracking_results}. \locil* outperforms both baseline models by a large margin. At this point, allow us to emphasize that this precision is remarkable seeing that \locil* was never informed about the location or existence of objects. Importantly, $96.6\%$ of slots that were recruited before the occlusion phase achieved a final tracking error (i.e., the tracking error in the moment the objects exit the scene) smaller than $10\%$, indicating that these slots tracked their self-assigned objects successfully throughout the entire scene. The poor tracking results of Loci-Unlooped and Loci-Visibility suggest that the internal loop and its adaptive control is critical for successfully tracking objects through occlusions. Please see Appendix~\ref{app:illustrations} for detailed illustrations of the scene and the corresponding slot representations.

The results furthermore show that \locil* learned to close the percept gates during occlusion, thus switching to a latent imagination mode (mean inner-loop integration: 99.2\%). Similarly, we find that the model made only sparse usage of observations when objects were visible (mean inner-loop integration: 91.1\%). 
Due to the percept gate opening regularization, \locil* has the inductive bias to predict the visible world while only glimpsing at it. 
As a result, the model trains itself on simulated occlusions besides the encountered ones during training, effectively generating further error signals that encourage the learning of more accurate latent temporally predictive models. 
This facilitates the generalization 
to extended occlusion scenarios. 

\begin{table}[h]
  \caption{Tracking results.}
  \label{table:tracking_results}
  \centering

   \resizebox{\textwidth}{!}{\begin{minipage}{1.1\textwidth}
  
    \begin{tabular}{lccc}
    \toprule
    
    & Mean & Successful & \multicolumn{1}{c}{MOTA} \\ 
    
    Model & Tracking Error & Trackings (\%) & \\

    \midrule

    \locil*  & $\mathbf{2.6}$ \small{$\pm$ 2.7} & $\mathbf{96.6}$ & $\mathbf{0.84}$ \\

    Loci-Visibility  &7.7 \small{$\pm$ 10.6}  & 43.6 & 0.64 \\

    Loci-Unlooped  &12.4 \small{$\pm$ 14.8}  & 7.4 & 0.76 \\

    \loci*  &12.5 \small{$\pm$ 10.3}  & 38.4 & -1.34 \\

    G-SWM  &26.8 \small{$\pm$ 14.5}& 7.1 & 0.23 \\
    
    SAVi  &26.7 \small{$\pm$ 12.6}& 3.2 & -0.67 \\
    
    \bottomrule
    \end{tabular}

  \end{minipage}}
\end{table}

\subsection{Violation of Expectation}
Having seen that \locil* tracks objects successfully through occlusions, we now test whether it has also learned to anticipate their reappearance.

\paragraph{Surprise scenario}

We focus on the ADEPT's vanish scenario that tests the concept of object permanence and directional inertia. The surprise condition (11 videos) features two objects that again traverse the scene behind the occluder this time, however, only one object reappears from behind the screen whereas the other vanishes while behind the screen. In the control condition both objects reappear. This scenario is designed to test the model's anticipation about the reappearance of the occluded object.

\paragraph{Slot Error}
To quantify an object- and thus slot-specific surprise we compute a slot error as follows:
\begin{equation}
    E_k^t = \frac{\sum_{i,j}\bigl[ (I^{t+1} - \hat{R}^{t+1}) \odot \hat{M}_k^{t,v} \bigr]^2}{\sum_{i,j} \hat{M}_k^{t,v}} ,
\end{equation}
where the overall prediction error is simply masked by the visibility mask of slot $k$. In addition, we divide the error by the sum of the visibility mask values to make the error invariant to the size of the object. For the following analysis we only consider slots that represent non-occluder objects and that achieved a final tracking error smaller than 10\%.

\begin{figure}[h]
    \centering
    \includegraphics[width=0.99\linewidth]{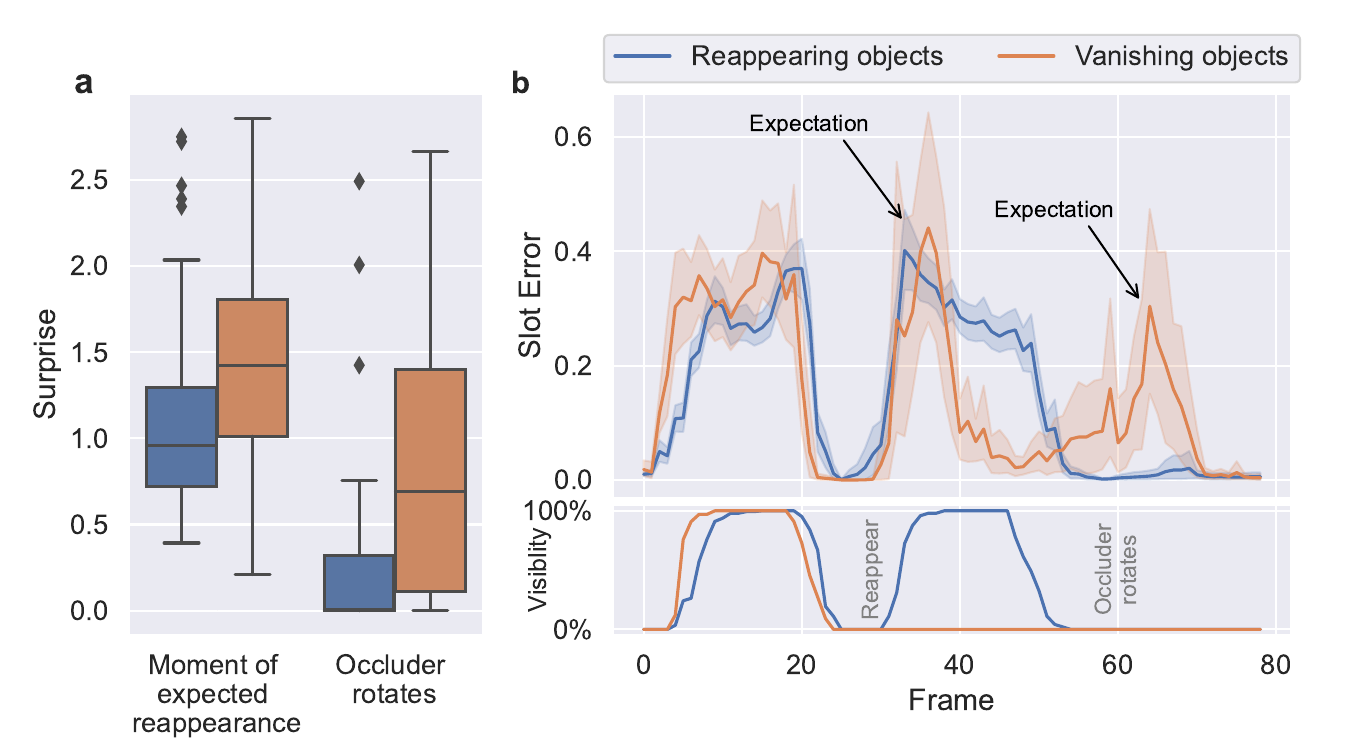}
    \caption{Results on the VoE experiment. Surprise is quantified as the maximum slot error in the corresponding frame interval.}
    \label{fig:voe}
\end{figure}

\paragraph{Results}
The model's surprise response indicates a significantly greater level of surprise when hidden objects fail to reappear showing a clear violation of expectation. Notably, this is the case for both time points: when the object should reappear after having slid past the occluder and when the occluder falls over after having not re-appeared before 
(cf., Figure~\ref{fig:voe}a; $t(75) = 1.69, p = .047; t(75) = 3.68, p < .001;$  as well as error peaks in Figure~\ref{fig:voe}b around frames 30 and 65). As shown in Figure~\ref{fig:controledemo},  \locil* tends to park the object behind the occluder if it did not reappear until the occluder falls over.
Note that this behavior is fully emergent, as \locil* is never trained on objects that permanently disappear behind occluders, and shows the model's strong bias to maintain stable, consistent object representations. 

\subsection{Sensory Interruptions}
Having seen that \locil* can handle the representation of partially observable scenes, we now investigate how it behaves when no observation is available for a brief period of time, simulating a short blink or visual blackout.

\paragraph{Dataset and Metric} 
The CLEVRER dataset \citep{yi_clevrer_2020} shows up to 6 small objects moving through a scene, including collisions and partial occlusions. We evaluate the next-frame prediction quality using PSNR, SSIM \citep{wang_image_2004}, LPIPS \citep{zhang_unreasonable_2018} and the segmentation quality using the Adjusted Rand
Index (ARI).

\paragraph{Sensory Interruptions}
In training and testing, we simulate sensory interruptions by setting the current input image to black with a probability of 20\%. In this case, the models need to maintain an internal stable scene representation without input information. They thus can only imagine how the scene will unfold. In the first 10 frames of each sequence we do not allow blackouts. 

\begin{figure}[t!]
    \centering
    \includegraphics[width=0.99\linewidth]{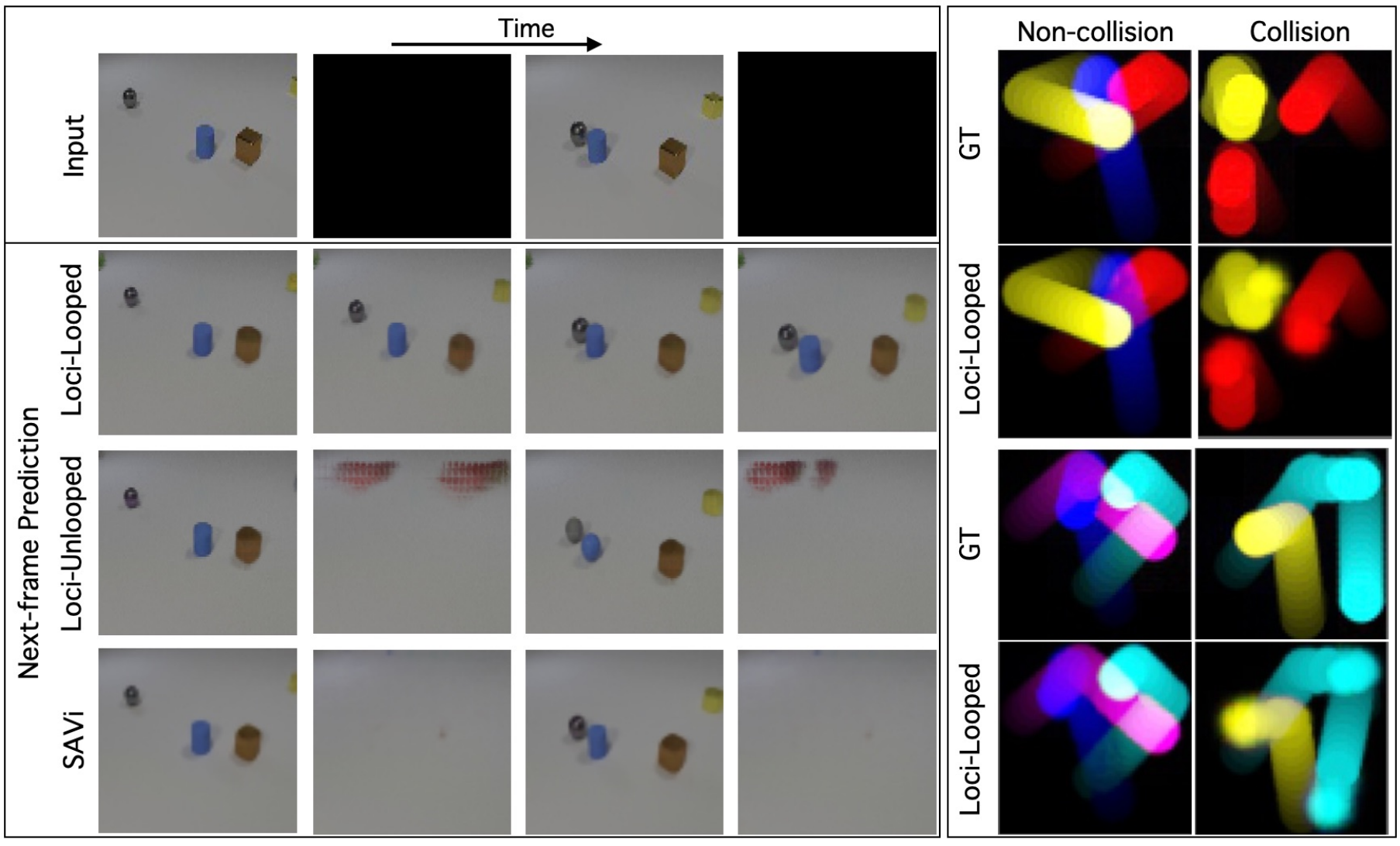}
    \caption{Left: Sensory interruptions exp. (CLEVRER dataset) Right: Imagination exp. (bouncing balls dataset). The trajectory of the first 20 predictions is shown.}
    \label{fig:clevrer}
\end{figure}

\paragraph{Results}
As depicted in Table~\ref{table:cvlevrer_results}, \locil* demonstrates superior performance compared to SAVi and Loci-Unlooped in timesteps with no available input frames and largely superior performance in timesteps with provided input frames. This observation implies that only \locil* can consistently uphold stable object representations during blackout periods, whereas the baseline models strongly depend on uninterrupted sensory input. The superiority over Loci-Visibility yet again confirms that the adaptive fusion gates integrate recurrent and sensory information highly effectively. Figure~\ref{fig:clevrer} shows qualitative results.

\begin{table}[t]
  \caption{Sensory interruptions results (CLEVRER dataset)}
  \vspace{0.1cm}
  \label{table:cvlevrer_results}
  \centering
    \resizebox{\textwidth}{!}{\begin{minipage}{1.2\textwidth}
    \hspace{0.2cm}
    \begin{tabular}{llcccc}
    
    \toprule
    Input &Method & PSNR & SSIM & LPIPS $\downarrow$ & ARI\\
    \midrule

    \multirow{4}*{Blackout} &\locil*    & $\mathbf{34.6}$ & $\mathbf{0.95}$ & $\mathbf{0.11}$   & $\mathbf{0.86}$\\
    &Loci-Visibility  & 30.4 & 0.92 & 0.18 & 0.74 \\
    &Loci-Unlooped     & 21.8 & 0.71 & 0.47 & 0.0 \\  
    & SAVi & 25.3 & 0.81 & 0.47 &  0.0 \\
    
    \midrule
    \multirow{4}*{Visible} 
    &\locil*    & 36.3 & $\mathbf{0.97}$ & $\mathbf{0.10}$   & $\mathbf{0.88}$ \\
    &Loci-Visibility  & 32.1 & 0.94 & 0.15 &  0.81 \\
    &Loci-Unlooped     & 28.1 & 0.86 & 0.26 &  0.46 \\
    & SAVi & $\mathbf{37.5}$ & 0.96 & 0.19 & 0.57 \\
    
    \bottomrule
    \end{tabular}

  \end{minipage}}
\end{table}

\subsection{Imaginations over long sequences}
While the above addressed sensory interruptions last only for short time periods,
we now test how well the model can generalize to long interruptions, by predicting scene dynamics over a long temporal horizon.

\paragraph{Dataset}
We use the bouncing balls dataset provided by \citet{lin_gswm_2020}, which shows three balls of different colors moving around. In the collision scenario, the balls collide with each other, while in the non-collision scenario, the balls are placed on different planes, and thus do not collide. 

\paragraph{Sensory Interruptions}
While training \locil*, we simulate sensory interruptions by setting the current input image to zero with a probability that increases linearly from 10\% to 45\% over the course of training. We provide the first 10 input images and then roll out the model thereafter. Importantly, during rollouts, while the base models G-SWM, STOVE \cite{Kossen2020Structured}, and SCALOR \cite{Jiang*2020SCALOR:} received their last prediction as input, Loci-Looped only receives sensory interruptions. 

\paragraph{Results} 

We find that \locil* is able to produce extended scene rollouts via the inner loop. In terms of accuracy, \locil* outperforms the baseline models in the collision scenario (see Table~\ref{table:bb_results}). The results suggest that \locil* has indeed developed a robust object-centric latent prediction model that can imagine complex object interactions over long time horizons.

\begin{table}
  \caption{Video imagination results (Bouncing balls dataset). Distance error of predicted ball positions, summed over the first 10 timesteps of generation. \footnotemark}
  \vspace{0.1cm}
  \label{table:bb_results}
  \centering
    \resizebox{\textwidth}{!}{\begin{minipage}{1.2\textwidth}
    \hspace{1.5cm}
    \begin{tabular}{lcc}
    \toprule
    Method  & Collision                               & Non-collision \\
    \midrule
    Loci-Looped      & $\mathbf{0.17} \pm 0.01$     & $0.20 \pm 0.01$                \\
    G-SWM           &  $0 . 2 4 \pm 0 . 0 3$        & $\mathbf{0 . 0 7} \pm 0 . 0 2$   \\
    STOVE           & $0.41 \pm 0.01$               & $0.39 \pm 0.02$                   \\
    SCALOR          &    $1.00 \pm 0.02$            & $0.27 \pm 0.02$                   \\
    \bottomrule
    \end{tabular}
  \end{minipage}}
\end{table}

\footnotetext{Results taken from \cite{lin_gswm_2020}}

\section{Discussion}
In this work, we introduced \locil*: an object-centric world model that has the ability to flexibly fuse outer loop sensations with inner loop imaginations. \locil* tracks objects through occlusion, learns the physical concepts of object permanence and directional inertia, and is robust to interruptions in its sensory signal. 
It builds on the idea that a holistic scene percept can be assembled object-wise from observations and imaginations. 
Importantly, and in contrast to competitive state of the art models, all of this was learned without supervision, without supervised slot initialization, without access to a temporal buffer; solely from the next-frame prediction objective. In line with \citet{piloto_intuitive_2022}, our work confirms that intuitive physics can emerge from learning an anticipatory world model that constantly predicts future world states. 

Future advancements of \locil* should incorporate probabilistic scene representations, particularly when multiple event continuations are possible \citep{smith_modeling_2019,gumbsch2023learning}.
This is especially the case for scenarios in which the agent acts while occluded---a typical feature of more advanced VoE scenarios. 
Furthermore, learning other and more complex object interactions, such as collisions, in a history-compressing architecture, such as the introduced \locil*, should be examined in further detail. 
Another limitation is the yet very simple nature of the considered datasets.
Recent approaches, including another Loci variant \cite{elsayed_savi_2022, traub_loci-segmented_2023, seitzer_bridging_2023}, suggest that bottleneck approaches paired with object pre-training are well-suited to handle real-world scenarios. 
An according application of \locil* is pending. 

Overall, we hope that the presented algorithms will contribute to further advance the development of more human-like compositional scene representation learning. 




\bibliography{bibliography}
\bibliographystyle{icml2024}

\clearpage
\newpage

\appendix
\section{Appendix}

\subsection{Extended Related Work}
\label{app:related}
Recently, two studies in the field of intuitive physics have gained attention for introducing a VoE dataset and models that learn the concepts of permanence, solidity, and continuity. Similiar to \locil*, the Physics Learning through Auto-encoding and Tracking Objects (PLATO) model \citep{piloto_intuitive_2022} uses a slot-wise encoder-predictor architecture. The second model is the Approximate Derendering Extended Physics and Tracking (ADEPT) model \citep{smith_modeling_2019} which implements a hand-crafted physical reasoning system. In this section, we will review both approaches and compare them with \locil*.

\textbf{Representation} \locil*, PLATO and ADEPT model physics at the level of objects. To incorporate this object-centric approach all models make use of a slot architecture, where a slot represents a processing pipeline dedicated to a single object. 
This slot-based architecture enables the parallel processing of multiple objects, applying the same model by weight sharing.
The models differ in their latent code constraints.
While PLATO does not constrain the latent code at all, ADEPT explicitly encodes the object's type, location, velocity, rotation, scale and color.
As \citet{smith_modeling_2019} has shown, this abstract encoding is beneficial for generalising to unseen objects, but requires supervised training. Balancing both approaches and inspired by the dorsal and ventral visual processing stream in humans, \locil* disentangles an object's position (where) and gestalt (what).

\textbf{Segmentation} To identify objects in an image PLATO relies on ground-truth segmentation masks, while ADEPT uses a supervised segmentation network. Unsupervised methods for image segmentation typically learn to decompose scenes into object-centric representations using slot-wise autoencoders \citep{burgess_monet_2019, greff_multi-object_2020}. Similarly, \locil* learns to identify objects in a scene using a slot-wise encoder-decoder architecture. By encoding positional information explicitly and constraining the gestalt encoding capacity, each slot is naturally biased towards representing a cohesive and uniform area of the image. While \loci* is capable of segmenting scenes with complex backgrounds using an additional high-capacity background slot \citep{traub_learning_2023}, this feature requires intensive training on the background. Our work focuses on short scenes with varying backgrounds, making it necessary to provide the model with the background for each scene.

\textbf{Dynamics Modelling} All three models leverage a dynamics module to estimate the state of the objects at the next timestep. While ADEPT does not learn objects dynamics but utilizes an out-of-the-box physics engine for this purpose, \locil* and PLATO make use of recurrent units. In PLATO, a slot-wise LSTM is combined with two feedforward networks accounting for pairwise object interactions to model object dynamics. \locil* differs with respect to the choice of the recurrent unit and how interactions are modelled. \locil* uses a slot-wise GateLORD \citep{gumbsch_sparsely_2022} module that penalizes latent state changes and thereby fosters stable hidden object state representations over time, while interactions between objects is modelled using multi-head self-attention between slots. 

\textbf{Tracking} Accurately identifying objects over time is crucial for estimating and predicting object motion. In practice, this means that recurrent prediction slots must receive consistent information about the same object over time to enable reliable predictions. To achieve this, PLATO relies on ground-truth information, while ADEPT utilizes a hand-crafted observation model that matches objects in the current observation with objects in the model's belief based on extracted object features. Similar, recent research has proposed an alignment module that learns to match object encodings between observations and a memory \citep{creswell_unsupervised_2021}. A different approach is taken by \locil*. 
The encoder module learns to consistently parse the same object in the same slot via a predictive coding approach, which yield the to-be-minimized reconstruction error of the previous time step as additional input.
Moreover, each slot of the encoder receives its previous output, thus priming its particular object-encoding responsibility. 
Finally, the internal GateLORD units as well as a time persistence loss further encourage latent encodings of the same object properties in the same slot over time. 

\textbf{Temporal memory} To predict the next position of objects, the models have to consider their movements. To do so, ADEPT's supervised perception module receives a history of three images and derives object velocities from it. In contrast, the recurrence in the dynamics modules of PLATO and \locil* allows to accumulate information over time and thus to capture object dynamics in the cell states. Although theoretically not needed, PLATO makes its prediction based on all past object encodings stored in an object buffer which is also used to derive object interactions. On the other hand, \locil*'s dynamic module predicts the next object state based solely on the current object state, requiring it to fully capture object dynamics within the current cell state.

\textbf{Object permanence} The ADEPT model does not learn object permanence which is by default built into the physics engine. In contrast, PLATO learns to predict the reappearance of hidden objects, which is however favored by access to the full history of object codes, informing about the previous existence of the object, and by a relative short duration of occlusion.

\subsection{Choice of Baseline Models}
\label{app:sec:baselines}

To our knowledge SAVi \citep{kipf_conditional_2022} is the most structural similar state-of-the-art model to \locil*. Both share the idea of an encoder-predictor-decoder architecture, and differ in their architectural details and overall inductive biases. G-SWM \cite{lin_gswm_2020} advocates its latent prediction kernel enhancing generative imaginations similarly to Loci-Looped. 

We did not run baseline comparisons against three recent powerful frameworks, namely SAVi++ \citep{elsayed_savi_2022}, Slotformer \citep{wu_slotformer_2023} and STEVE \citep{singh_simple_2022}. In our understanding these comparisons are of limited use. SAVi++ main extension is its improved performance on real-world datasets, incorporating camera motion and explicitly exploiting ground-truth depth information in training. Neither of these characteristics apply to our study of object permanence and our datasets. Moreover there is no architectural improvement from SAVi to SAVi++ that would address the problem of maintaining stable slot representations of temporarily hidden objects, suggesting that the performance of SAVi is a good indicator on how SAVi++ will perform on our tests. Slotformer, on the other hand, is not a compositional scene representation model but a slot-based video prediction model that trains and relies on pre-computed slot-representations, for example, computed using SAVi or STEVE. This dependency makes a comparison with Slotformer not very informative as the model's task is different. We were also not able to provide baseline results for STEVE because its transformer based decoder does not allow to decode slot representations into object masks, instead slot attention masks of the input are used to visualise slots. This however implies that STEVE cannot produce masks for hidden objects which we evaluated in our experiments.

Concerning the intuitive physics models: We were not able to train PLATO \citep{piloto_intuitive_2022} on the ADEPT vanish scenario (as also stated in \citet{piloto_intuitive_2022}), because the model expects aligned input masks that need to be provided consistently. In addition, PLATO requires a very coarse temporal resolution (15 frames for one video) simulating only short occlusions, whereas Loci-Looped and SAVi can be trained on fine temporal resolutions (41 frames) simulating longer occlusions. We did not include the ADEPT \citep{smith_modeling_2019} model as baseline as it would be a skewed comparison in our opinion. The model depends on supervised information to train its encoder, its decoder and its particle filter. 
Moreover, the ADEPT model uses an out-of-the-box physics engine. We did not include baselines without explicit object representations as numerous related work suggest that object agnostic models perform inferior \citep{piloto_intuitive_2022, smith_modeling_2019,  wu_slotformer_2023}.

\subsection{Ablation Studies}
\label{app:ablations}
\begin{figure}[h!]
    \centering
    \includegraphics[width=0.99\linewidth]{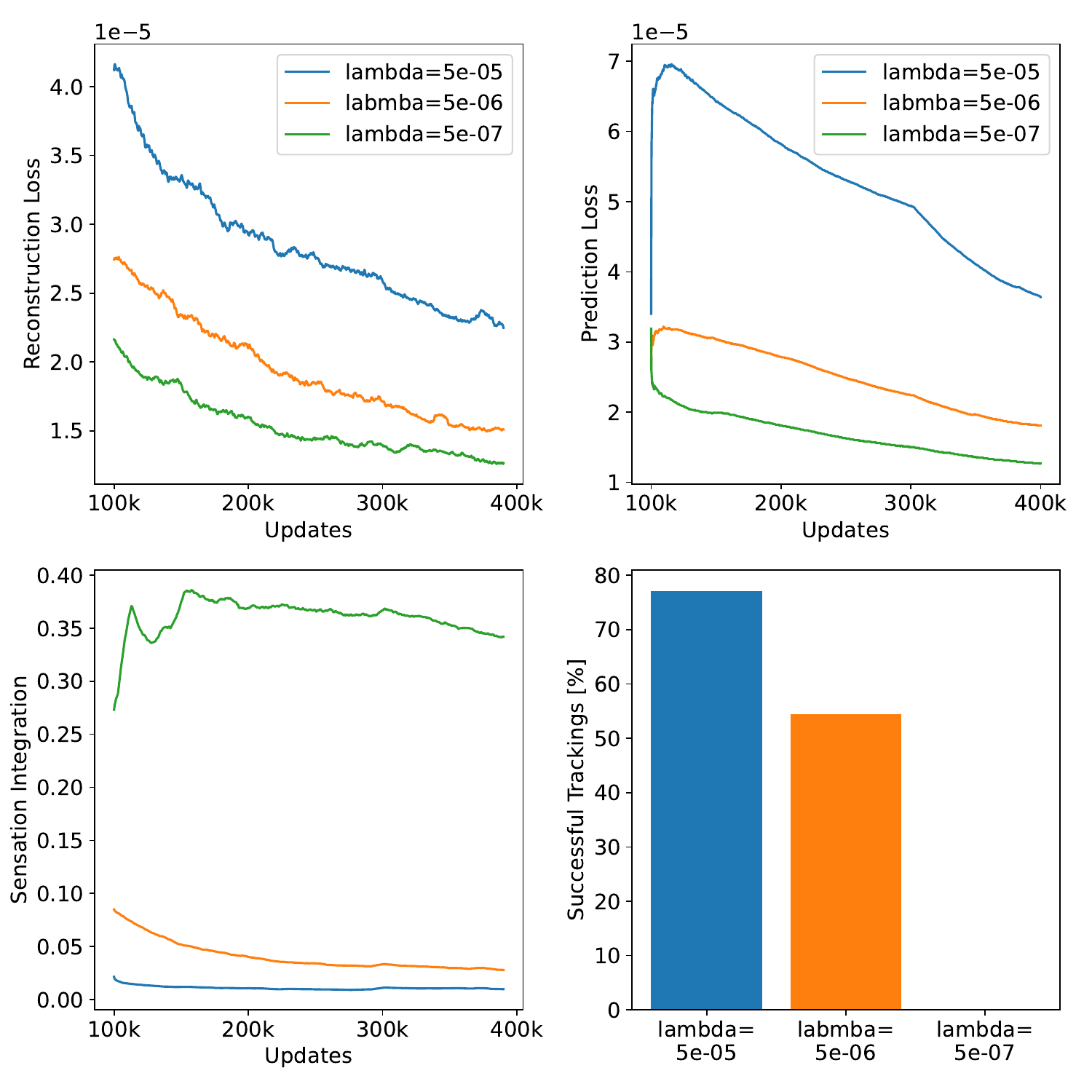}
    \caption{Results for the ablation of the Percept Gate Opening Regularization weight lambda.}
    \label{fig:app:ablation_lambda}
\end{figure}

We have ablated the weight ($\lambda$; see~\ref{app:updategate}) of the Percept Gate Opening Regularization term by running experiments on the ADEPT dataset for 300k updates. A larger lambda corresponds to a larger penalty on the integration of sensations (i.e., the current observation from the encoder) to form the current percept. That is, the smaller lambda the smaller the incentive for the model to use the inner loop and to rely on its own predictions. As illustrated in Figure~\ref{fig:app:ablation_lambda}, we found that high regularization leads to a less accurate (pixel-wise) reconstruction of the current frame and a less accurate (pixel-wise) prediction of the next frame. 
This indicates that encouraging the usage of the inner loop has a slightly negative effect on learning accurate reconstructions and next step predictions. 
This effect may cease, though, when the system has even more time to learn. 
On the other hand, we observe that the usage of the inner loop enforced by a larger lambda is crucial for learning to track objects through occlusions and for learning the concept of object permanence---most likely because the loss encourages a closed inner loop, and thus a larger gradient flow towards the inner loop. Consequently, the model with the highest regularization achieved the best tracking results at test time (see table~\ref{table:ablation_results}).

Concerning the reconstruction loss we found that omitting the term has observable negative effects on learning (see Figure~\ref{fig:app:ablation_recon}). Our ablation study suggests that the final performance of the model improves with a stronger weighting of the reconstruction loss. This is most likely because the perceptual codes become less accurate in encoding the current input frame. In addition, we experienced that the reconstruction loss stabilizes the model training.

\begin{table}[h]
  \centering
  \caption{Ablation results on the weighting of the precept gate openings regularization (Lambda) and the weighting of the input-reconstruction loss ($L_{Rec}$).}
  \label{table:ablation_results}

   \resizebox{\textwidth}{!}{\begin{minipage}{1.15\textwidth}
  
    \begin{tabular}{lcccc}
    \toprule
    
    & \multicolumn{2}{c}{Mean} & Successful & \multicolumn{1}{c}{MOTA}\\ 
    
    &\multicolumn{2}{c}{Tracking Error (\%)} & trackings (\%) &\\
    
    \cmidrule(r){2-3}\cmidrule(r){4-4}\cmidrule(r){5-5}
    
    Model & Visible & Occluded & Overall & Overall\\
    
    \midrule

    Lambda=5e-05  & $\mathbf{3.3}$ & $\mathbf{2.8}$ & $\mathbf{77.1}$ & $\mathbf{0.65}$ \\
        
    Lambda=5e-06  &5.3 & 5.2  & 54.5 & 0.60\\ 

    Lambda=5e-07  &11.6  & 7.1  & 0 & 0.61\\
    
    \midrule

    $L_{Rec} = 0.33$  & $\mathbf{3.2}$  & 3.5 & $\mathbf{96.0}$ & 0.65 \\

    $L_{Rec} = 0.1$  &3.3 & $\mathbf{3.3}$ & 84.5 & $\mathbf{0.67}$ \\
            
    $L_{Rec} = 0.0$  &4.6 & 3.9 & 70.5 & 0.60\\
    
    \bottomrule
    \end{tabular}
  \end{minipage}}
\end{table}

\begin{figure}[h!]
    \centering
    \includegraphics[width=0.99\linewidth]{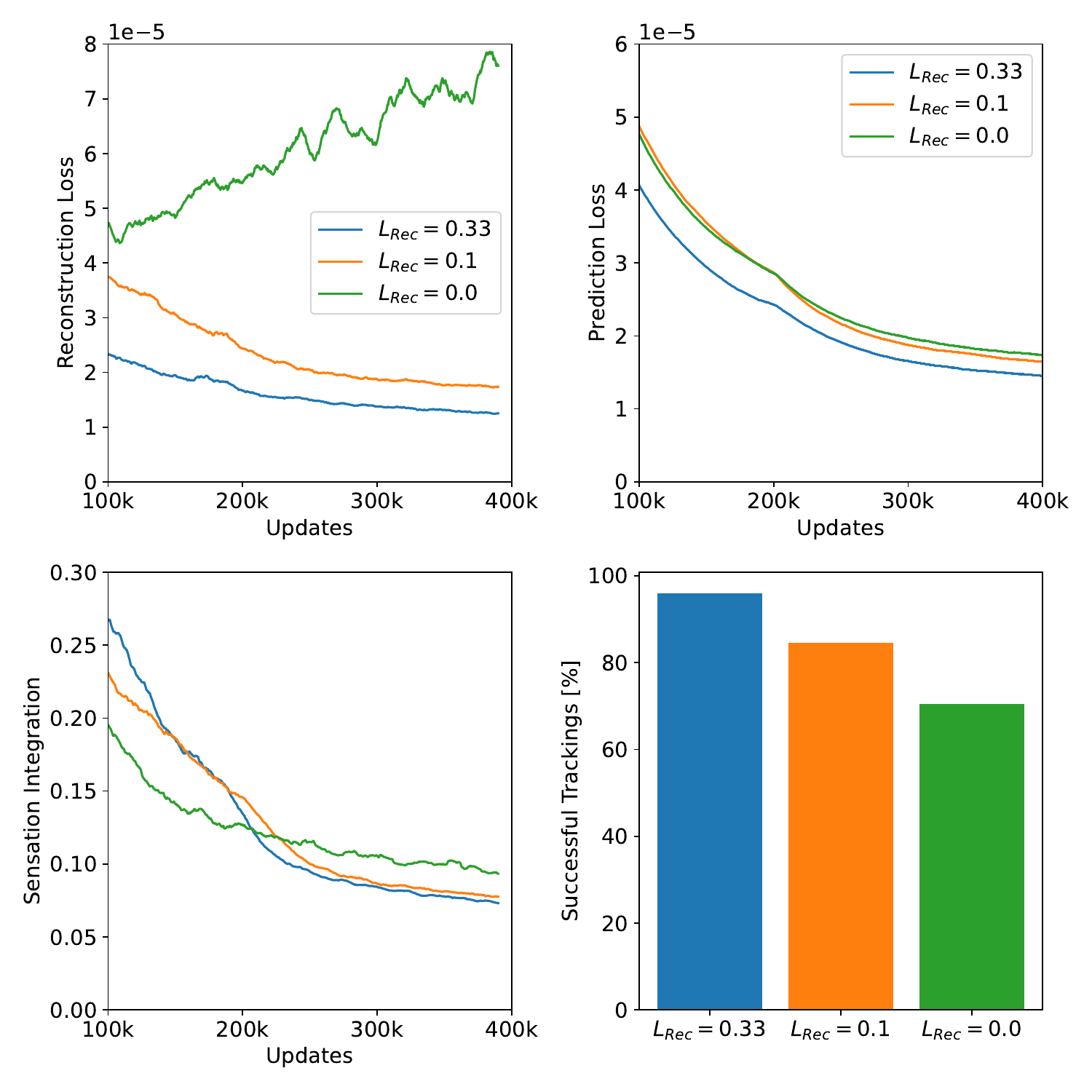}
    \caption{Results for the ablation of the Reconstruction Loss.}
    \label{fig:app:ablation_recon}
\end{figure}

\begin{figure}[h!]
    \centering
    \includegraphics[width=0.99\linewidth]{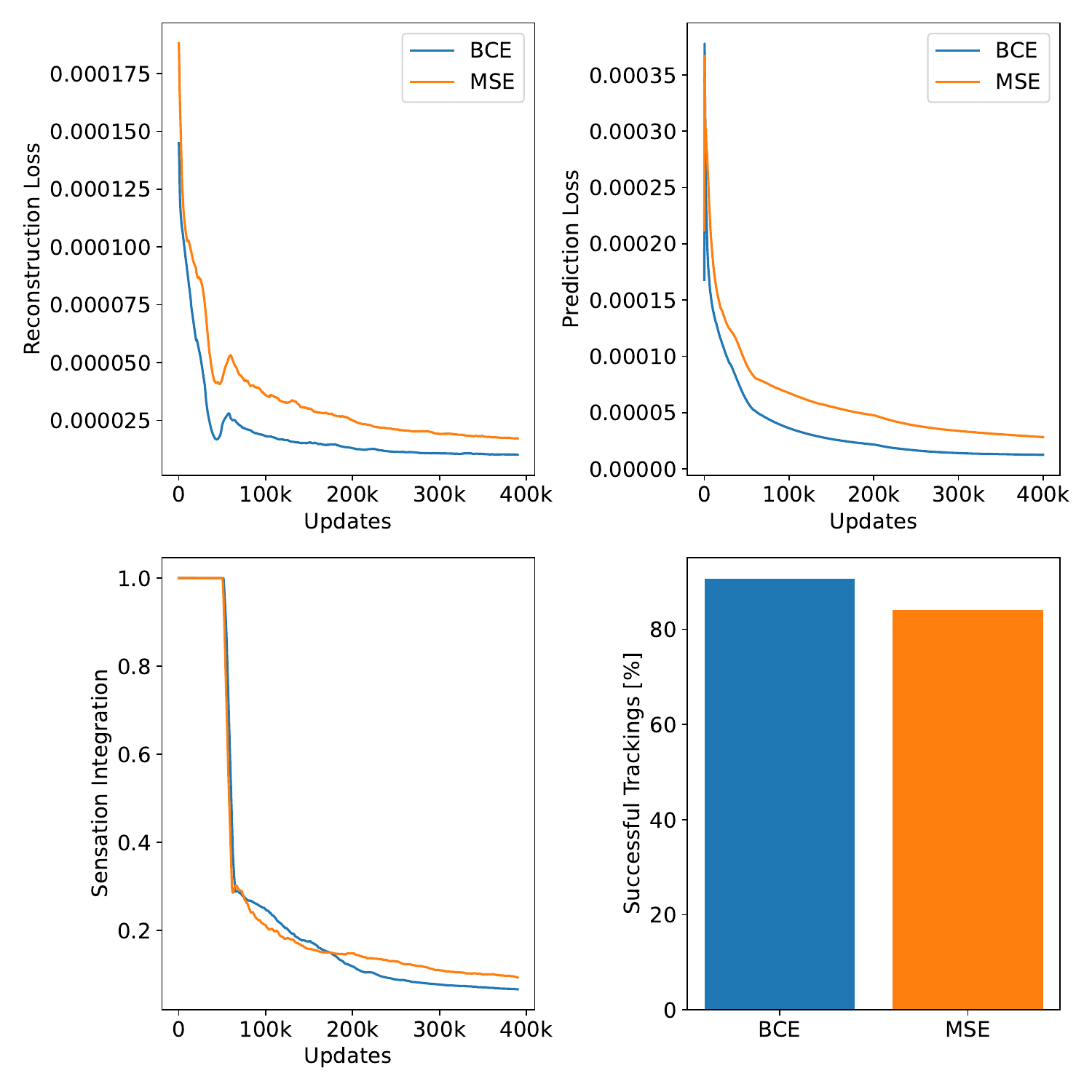}
    \caption{Results for the ablation of the image loss term used in the input-frame reconstruction loss and next-frame prediction loss.}
    \label{fig:app:ablation_recon}
\end{figure}

\newpage
\subsection{Method}
\label{app:method}


\subsubsection{Percept Gate Controller}
\label{app:updategate}
The percept gate controller is part of the slot-wise gate module and computes $\alpha_k^{t,G}$ and $\alpha_k^{t,P}$. The controller receives the inputs $\tilde{P}^t_k$, $\tilde{G}^t_k$, $\tilde{O}^t_k$, $\hat{P}^t_k$, $\hat{G}^t_k$, $\hat{O}^t_k$, $\hat{P}^{t-1}_k$ which are concatenated into a vector of size 206. This vector is then fed into a feed-forward network modelling update function $g_{\theta}$. 
This network is composed of three linear layers with dimensions 32, 16, and 2, and employs the hyperbolic tangent activation function. During the training phase, the network's output is augmented with Gaussian noise ($\Sigma = 0.1$); however, this is not applicable during the inference stage. As demonstrated by \citet{gumbsch_sparsely_2022}, the stochastic component fosters the learning of sparse gate openings. The final values for $\alpha^{t,G}_k$ and $\alpha^{t,P}_k$ are calculated using the rectified hyperbolic tangent function ($\Lambda$). The rectified variant generates values within the range $[0,1)$, thus enabling the model to fully close the gates (i.e., $\alpha = 0$). In the backward propagation, the following pseudo-derivative is employed:
\begin{equation}
    \Lambda' = \frac{\partial \Lambda(x)}{\partial x} = \begin{cases} 0 & \text{if } x \le 0\\ (1 - \Lambda(x)^2) & \text{otherwise} \end{cases} 
\end{equation}
In addition, we penalize gate openings (i.e. $\alpha > 0$) by applying a $L_0$ regularization. We therefore use the method described in \citet{gumbsch_sparsely_2022}. The regularization loss is given as the sum of gate openings: 
\begin{equation}
   L_{\textrm{Gate}} = \lambda \sum_k (\Theta(\alpha^{t,G}_k) + \Theta(\alpha^{t,P}_k)), 
\end{equation}
where weighting factor $\lambda$ controls the strength of the loss and $ \Theta$ is the non-differentiable Heavisite step function. We therefore use the derivative of the linear function as the pseudo-derivative:
\begin{equation}
    \Theta' = \frac{\partial \Theta(x)}{\partial x} = 1.
\end{equation}

\subsubsection{Slot Recruiting}
\label{app:slotrecuiting}
A crucial step in slot-based architectures is the initial assignment of slots to objects. \citet{traub_learning_2023} demonstrated that \loci* can allocate multiple slots in parallel to identify multiple objects. However, this allocation scheme can be sensitive to object sizes. Specifically, large objects are more complicated to reconstruct than small objects and thus provoke larger prediction errors. As a consequence large objects tend to attract multiple slots in parallel, resulting in one object being encoded partly in multiple slots. To encourage the representation of entire objects in exclusively one slot, we restrict the encoder to only use one slot at a time seeking for new objects. In general, we distinguish between occupied slots which already represent an object and empty slots which do not represent an object yet. Once both the visibility masks $\tilde{M}_k^{t+1,v}$ and $\hat{M}_k^{t+1,v}$ exceed a threshold of 0.8 in one pixel, the corresponding slot is marked as occupied for the entire sequence. At the start of a sequence \locil* has one empty slot available. When this empty slot becomes occupied a new one is recruited with a delay of two timesteps. This delay allows the initial slot to encode the object in its entirety. This pattern repeats when the empty slot becomes occupied again. In addition, every second frame the isotropic Gaussian $\hat{Q}^t_k$ of the empty slot is set to the position $(x,y)$ of the largest prediction error in the background i.e. 
\begin{equation}
    (x,y) = \operatorname*{argmax}_{(i,j)} \ (M_{bg} \odot E)(i,j)
\end{equation}
before entering the encoding process. With this incremental recruiting scheme, \locil* encodes entire objects of varying sizes more reliably in one slot.

\subsubsection{Training procedure}
\label{app:trainingprocedure}
The training procedure entails randomly selecting sequences from the dataset and compiling them into a single batch. This batch is then processed sequentially, with the model ingesting consecutive frames and executing a backward pass every $n$ frames. Simultaneously, an optimizer step is conducted every $n$ frames, followed by the detachment of gradients. Only the internal hidden states remain unaltered, and they are cleared only after the full batch of sequences has been processed. Similarly, the eprop eligibility traces employed within the GateL0rd layers are maintained for each sequence. It is important to highlight that these eligibility traces effectively facilitate the integration of error information from the past beyond the truncation horizon of backpropagation-through-time by accumulating previous neuron activations, akin to the approach described in \citet{bellec2019biologically} which facilities the the learning of long lasting memory states as previously demonstrated by \citet{traub_learning_2023}.

Training the model in an unsupervised fashion is challenging which requires increasing the difficulty of the task in three phases. During the first phase, the focus is on learning to represent foreground objects. Therefore, the reconstruction and the prediction loss are initially only applied to the foreground by masking the corresponding targets, 
\begin{align}
   I^{t\prime} & =  I^t \odot M^{t}_{fg} + I^t \odot (1-M^{t}_{fg}) \cdot \beta 
   \\
    M^{t}_{fg} & = \theta < (I^t - I^{t}_{bg})^2  ,
\end{align}
where $\beta$ is set to zero. To encourage initial slot bindings, all slots are placed in parallel and in a stochastic fashion to the largest foreground errors. By the end of the first phase, the model should be able to use the slots to rudimentarily reconstruct and predict the foreground. In the second training phase the aim is to learn to represent entire objects in one slot, for which slot recruiting (see Section \ref{app:slotrecuiting}) is enabled. In addition, the background is blended in the losses by gradually increasing $\beta$ to one. This enforces the learning of background mask $M_{bg}$ which is used to distinguish between background and foreground. At the end of phase two, the model should be able to reconstruct and predict complete scenes. Until this point, the update module was skipped focusing the training on the outer loop and visible objects. This is changed in the last training phase, in which the update module is enabled and the model's imagination is trained. \locil* then learns to balance information from the inner and the outer loop. 

\subsubsection{Teacher forcing}
Following \citep{traub_learning_2023}, \locil* starts a sequence by repeatedly processing the first frame $x$ times (teacher forcing phase). The prediction target is given by first frame as well. This allows the model to identify initial objects in the scene using slot recruiting. In this phase the updatemodule and transition module are skipped, basically using the encoder and decoder module as slot-wise auto-encoder for an initial scene segmentation.  

\subsubsection{Training specifics}
\label{app:hyperparameter}
From 200k updates on wards we summed the gradients over two timesteps and then ran one joint optimization step. In addition, we applied a dropout on the prediction error $E_t$ before entering the encoder ($p=0.1$). 

\begin{table*}[h]
  \caption{Loci Training Specifics}
  \label{table:params}
  \centering
  
  \begin{tabular}{lccc}
    \toprule
    Parameter                             & ADEPT                    & CLEVRER                  & Bouncingballs \\
    \midrule
    Learning Rate                         & $1 \cdot 10^{-4}$        & $1 \cdot 10^{-4}$        & $1 \cdot 10^{-4}$ \\
    Learning Rate (from 400k updates)     & $3.3 \cdot 10^{-5}$      & $3.3 \cdot 10^{-5}$      & - \\
    Batch size                            & 16                       & 32                       & 128 \\
    Number of updates                     & 1150000                  & 800000                   & 800000 \\
    Teacher forcing length                & 10                       & 10                       & 5 \\
    Resolution                            & 120x80                   & 120x80                   & 64x64 \\
    Resolution (from 600k updates)        & 480x320                  & 120x80                   & - \\
    Number slots (objects)                & 7                        & 6                        & 3 \\
    Start training phase 2                & 30k updates             & 30k updates             & 15k updates \\
    Start training phase 3                & 60k updates             & 60k updates             & 30k updates \\   
    GateLORD Regularization               & $1 \cdot 10^{-10}$      & $1 \cdot 10^{-10}$      & $1 \cdot 10^{-10}$ \\
    Video length (training)               & 41 frames               & 64 frames               & 100 frames \\ 
    Training set size                     & 1000                     & 20000                    & 10000 \\        
    Frame offset                          & 3 frames                & 2 frames                & - \\ 
    $\lambda$                             & $5\cdot10^{-6}$         & $5\cdot10^{-6}$         & $1\cdot10^{-5}$ \\
    \bottomrule
  \end{tabular}
\end{table*}

\subsubsection{SAVi Training}
For training SAVi \cite{kipf_conditional_2022}, we used the stochastic SAVi implementation as well as the hyper-parameters provided by \citet{wu_slotformer_2023}. We used a resolution of 64x64 for both the ADEPT and the CLEVRER dataset. For training efficiency (see \citet{wu_slotformer_2023}) we trained SAVi on subsequences of the full videos which had length 6. For the ADEPT dataset, we trained SAVi for 4 epochs, for longer training we observed that the model overfitted to the background and started to neglect foreground objects. For the CLEVRER dataset, we trained SAVi for 12 epochs including simulated blackouts in the training.

\subsubsection{G-SWM Training}
For training the G-SWM model we used the code provided by \citet{lin_gswm_2020}. We used a resolution of 64x64 for the ADEPT dataset and trained on full sequence length. We trained two models until convergence (250000 updates) using the hyperparameters provided in the repository for the similar OBJ3D dataset. To enable a fair comparison against \locil*, we provided the scene background as supervised input to the G-SWM model in training and testing. 

\subsubsection{Loci-v1 Training}
For training the Loci-v1 model we used the code provided by \citet{traub_learning_2023}. We used the same training procedure as for \locil* and the hyperparameters listed in Table~\ref{table:params}.

\subsection{Exp 1: Tracking}
\subsubsection{Training Set}
\label{app:exp1}
Each video of the ADEPT dataset contains a different background and a static camera perspective. We used 90\% of videos for training and 10\% for validation. In addition, we increased the video speed by considering only every third frame, which gives a video length of 41 frames. We trained 3 independent models of \locil* for the ADEPT dataset and averaged the results across slots.

\subsubsection{Extracting object masks}
To obtain position estimates of the encoded slot objects in the SAVi and the GSW-M model we extracted self-computed object masks, as both models do not create object masks per default. For SAVi we extracted the slotmasks computed before the softmax operation and considered them as obejct masks. For G-SWM we used the individual object reconstruction masks ($\hat{o}$ and $\hat{\alpha}$) before they were multiplied with the presence variable $z_{pres}$. To translate the object masks into position estimates, we fitted a rectangular bounding box around the object mask and then calculated the euclidean center of it. These slot positions were then used in the tracking and MOTA evaluation. To determine whether a slot had been recruited to encode an object in the SAVi model, we applied a threshold (0.9) to the object mask. For the G-SWM model we thresholded the slot-specific presence variable $Z_{pres}$.

\subsubsection{Tracking Error}\label{app:tracking}
\locil* encodes object positions explicitly in 2D image coordinates which is highly interpretable, allowing us to easily quantify the model's tracking precision as the distance between estimated object positions and the true object positions. To do so, we pair the models internal representations with the ground-truth objects in the scene. More specifically, each time the model detects a new object the current positional encoding is used to assign the slot to the closest object in the scene based on euclidean distance. This pairing is then locked. Finally, at each timestep the slot-specific tracking error $T^t_k$ is computed as the euclidean distance between the estimated and the true object position. Lastly, we scale the error to the interval 0 to 1 by dividing the error by the image diagonal $d$:
\begin{equation}
 T^t_k = \frac{\sqrt{(\hat{P}_k^t - P_o^t)^2}}{d}    ,
\end{equation}
where $P_o^t$ is the true position of the assigned object.

\subsubsection{Multiple Object Tracking Accuracy}
In addition, we record the Multiple Object Tracking Accuracy (MOTA) \citep{bernardin_evaluating_2008} to quantify how well the model detects objects and identifies them temporally consistent. The MOTA is given as:
\begin{equation}
\mathrm{MOTA} = 1- \frac{\sum_t(\mathrm{FN}_t + \mathrm{FP}_t + \mathrm{IDS}_t)}{\sum_t\mathrm{GT}_t} ,
\end{equation}
where FN denotes the number of false negative detections, FP the number of false positive detections, IDS the number of object switches between slots, and GT the true object instances. Each timestep the ground-truth objects in the scene are paired with the slot representations based on euclidean distance (see Appendix~\ref{app:tracking}). As this is a one-to-one mapping, MOTA counts unassigned slots as false-positives and unassigned objects as false-negatives. Object switches occur if an object is first assigned to slot a and later to slot b. Importantly, we also provide the position of occluded objects as part of GT. 

We used the python package \textit{motmetrics} for computing the MOTA (https://github.com/cheind/py-motmetrics). Pairwise distances between slot positions and ground-truth positions were calculated using euclidean distance, where the cutoff distance was set to 10\% of the image diagonal. Further, we only considered occupied slots (see Section~\ref{app:slotrecuiting}) which, made a position estimate within the image borders, and predicted their slot-object to exist ($\sum_{(i,j)} (\hat{M}^{t,o}_k) > 100$, i.e. the object mask size exceeds a threshold of 100).

\subsubsection{Violation of Expectation}
\begin{figure}[h!]
    \centering
    \includegraphics[width=0.9\linewidth]{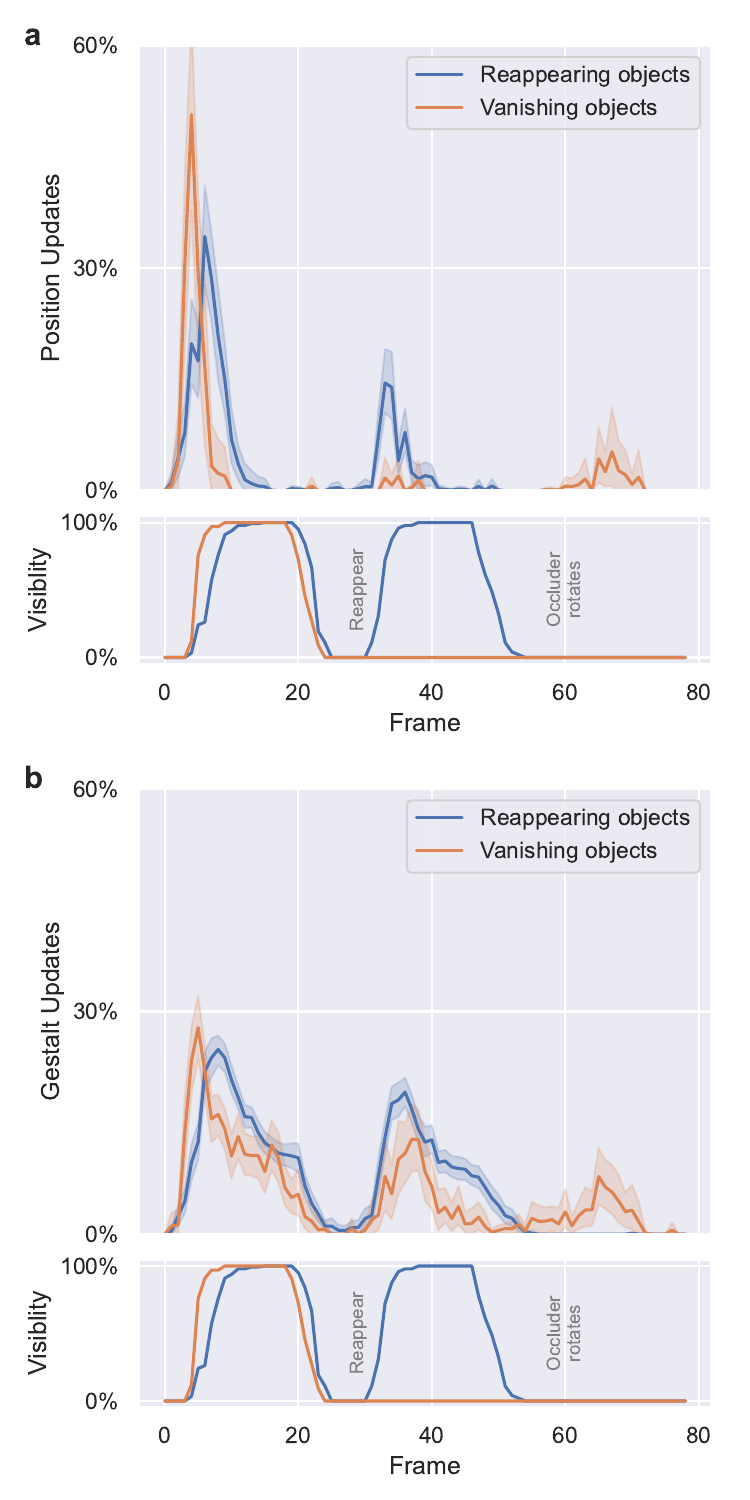}
    \caption{Gate openings on the Violation of Expectation experiment. Traversing objects either reappear from occlusion (blue) or vanish in occlusion (red). We observe that the position gates mainly open for sensory information after objects initially appear and after they reappear after occlusion. The gestalt gates in contrast integrate constantly sensory information if the objects are visible or expected to become visible.}
    \label{fig:gateopenings}
\end{figure}

Given the parallel trajectories of objects reappearing and vanishing, the anticipated moment of object reappearance correlates with the increased visibility of reappearing objects (observed post-frame 30 in Figure~\ref{fig:voe}b). At this juncture, a noticeable surge in slot error aligns precisely with the expected moment of object reappearance. Intriguingly, this surge is also evident for vanishing objects, suggesting the model's anticipation of their reappearance at this specific timepoint. 
This is confirmed by a significant correlation between the slot error of vanished objects and the visibility of reappearing objects (frames: 25-40, $r(13)=.9, p < .001$). Likewise, we find the same pattern for the size of the visibility mask (frames: 25-40, $r(13)=.94, p < .001$), indicating that \locil* expected the vanished objects to become visible again with the expected moment of reappearance. Interestingly, we find a second peak of expectation in the moment the screen flips to the ground, failing to reveal the missing object. Interestingly, similar observations can be made when monitoring the gate opening behavior \autoref{fig:gateopenings}.

To test the difference in surprise between the reappearing and the vanishing trials, we employ one-sided two-sample t-tests according to our hypothesis that the surprise is larger for vanishing objects than for reappearing objects. 

\newpage
\subsection{Exp 2: Sensory Interuptions}
For the CLEVRER Dataset we used the evaluation scripts provided by \citet{wu_slotformer_2023} as well as their training and testing split. Sensory interruptions were simulated by setting the current input frame $I^t$ and the previous prediction error $E^t$ to zero before providing them to the encoder. The encoder thus did not receive any new GT scene information. The model then predicts the next-frame for which the target is provided. Consequently, the next-frame prediction loss used for optimization is not affected and is calculated without interruption. 

\subsection{Exp 3: Video Prediction}
For the Bouncing Balls Dataset, we leveraged the dataset generation and evaluation scripts provided by \citet{lin_gswm_2020}. The baseline results, also sourced from \citet{lin_gswm_2020}, served as a reference. During training, we opted to enhance the effectiveness of gestalt change regularization by increasing its weighting to 0.25. During the testing phase, we supplied Loci-Looped with the initial 10 frames of the video sequence, followed by a prolonged sensory interruption. To estimate the positions of the balls, we deviated from using the internal positional code, as our observations indicated that Loci-Loped did not encode the object position at the center of the balls but rather at the top border of the shape. Instead, we implemented a strategy where we fitted a rectangular bounding box around the object mask, determined the center of this box, and utilized it as the positional estimate for the ball.


\subsection{Illustrations}
\label{app:illustrations}
In this section, we illustrate how the models effectively utilize their slots to represent the scenes employed in our experiments.

\begin{figure*}[h!]
    \centering
    \includegraphics[width=0.8\linewidth]{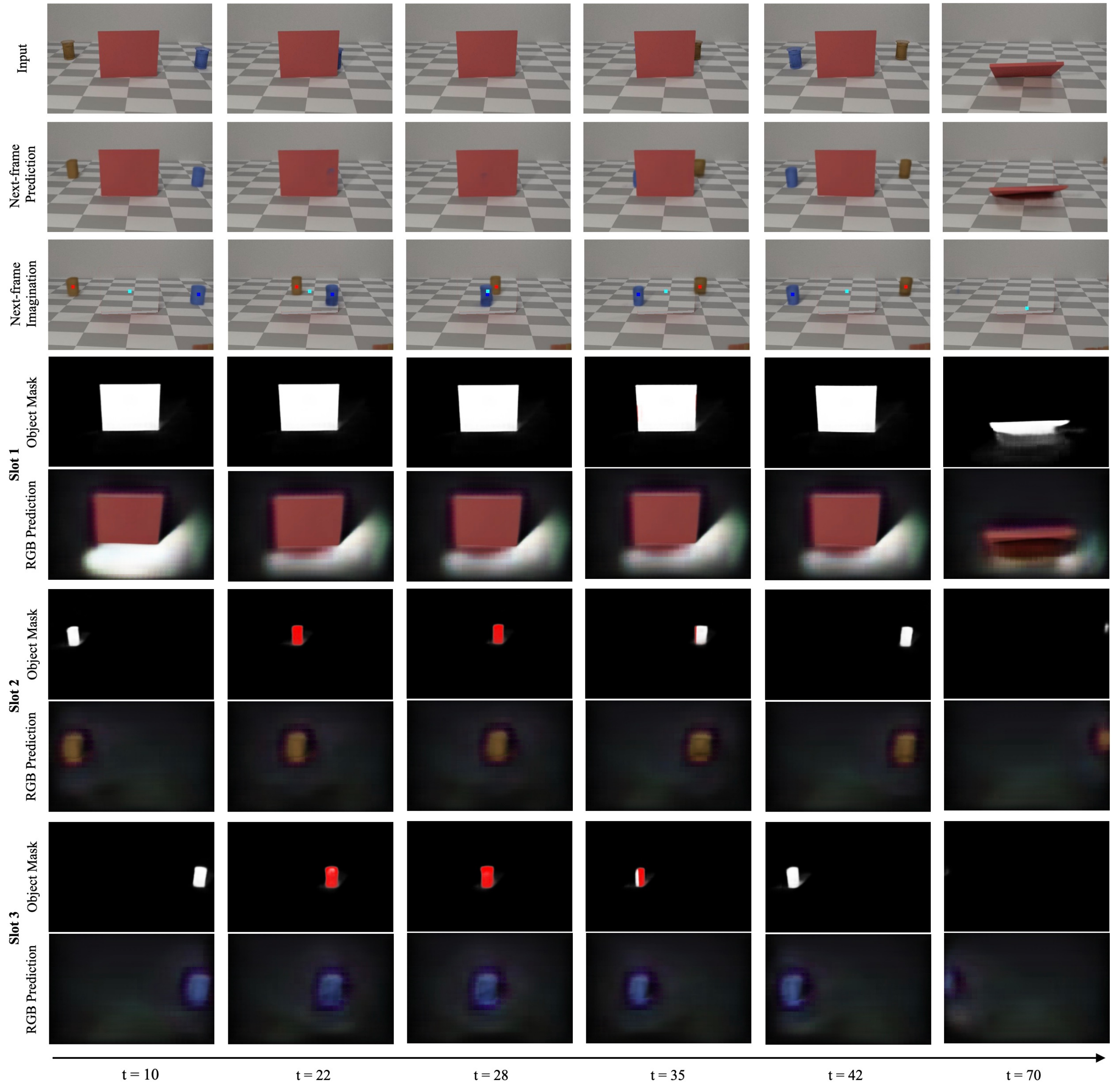}
    \caption{\textbf{The control sequence of the Violation of Expectation experiment and \locil*'s perception of it.} \locil* maintains clear object representations throughout the occlusion phase. \textit{Input:} The current frame of the sequence which serves as input. \textit{Next-frame prediction} \locil*'s composed RGB prediction for next timestep. \textit{Next-frame Imagination} \locil*'s composed RGB prediction for next timestep without the occluder screen. The colored dots illustrate the GT positions of the objects. \textit{Slot-wise object mask:} \locil*'s predicted object masks depict full object shapes. Red colored parts correspond to occluded object parts and white colored parts to visible object parts. \textit{Slot-wise RGB prediction:} \locil*'s predicted reconstruction of the object in pixel-space.}
    \label{fig:app:controldemo}
\end{figure*}

\begin{figure*}[h!]
    \centering
    \includegraphics[width=0.8\linewidth]{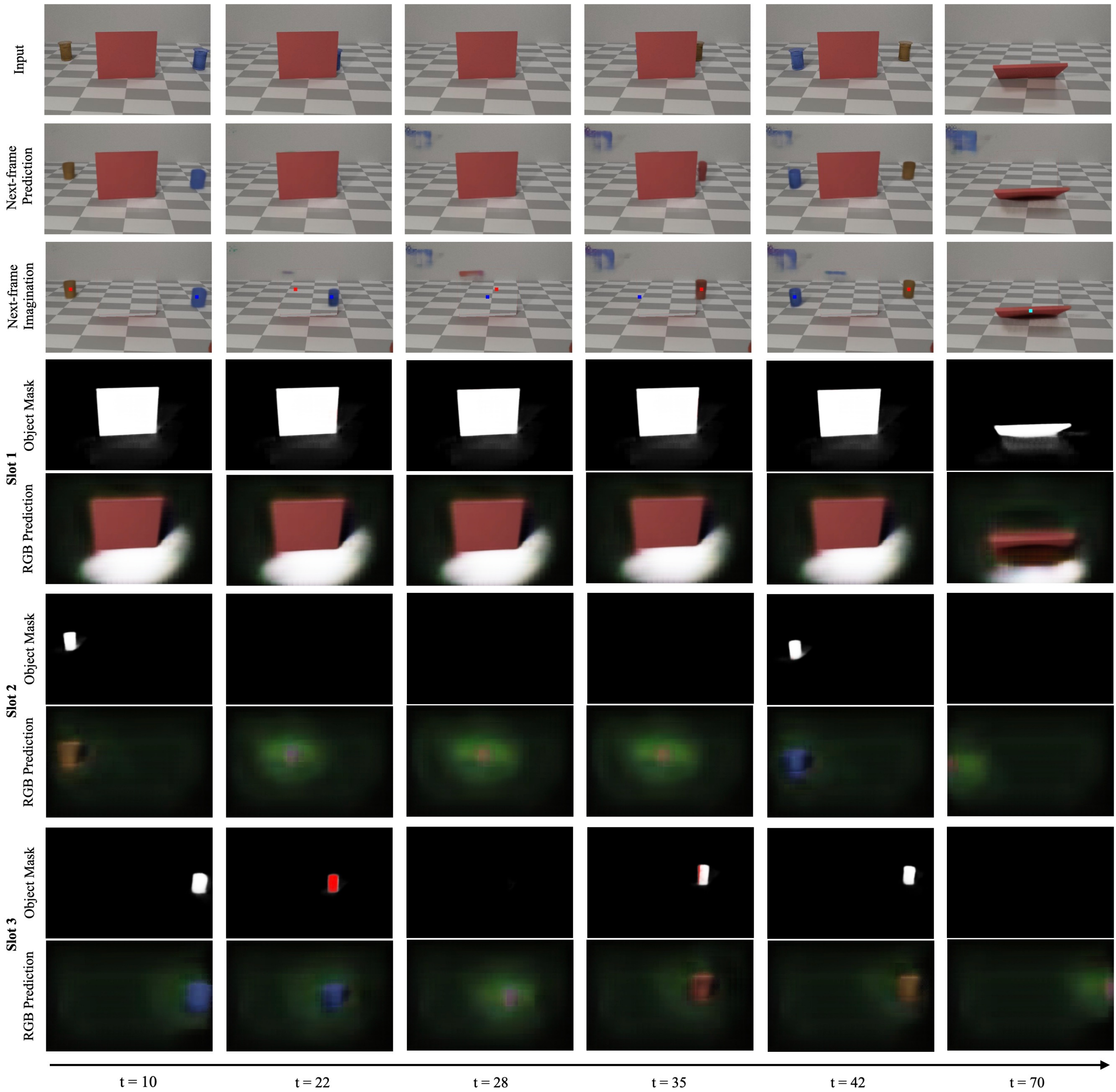}
    \caption{\textbf{The control sequence of the Violation of Expectation experiment and Loci-Unlooped's perception of it.} Loci-Unlooped does \textit{not} maintain clear object representations when objects become occluded. The reappearing objects are switching slots showing inconsistent tracking of temporarily hidden objects. \textit{Input:} The current frame of the sequence which serves as input. \textit{Next-frame prediction} Loci-Unlooped's composed RGB prediction for next timestep. \textit{Next-frame Imagination} Loci-Unlooped's composed RGB prediction for next timestep without the occluder screen. The colored dots illustrate the GT positions of the objects. \textit{Slot-wise object mask:} Loci-Unlooped's predicted object masks depict full object shapes. Red colored parts correspond to occluded object parts and white colored parts to visible object parts. \textit{Slot-wise RGB prediction:} Loci-Unlooped's predicted reconstruction of the object in pixel-space.}
    \label{fig:app:controldemo}
\end{figure*}

\begin{figure*}[h!]
    \centering
    \includegraphics[width=0.7\linewidth]{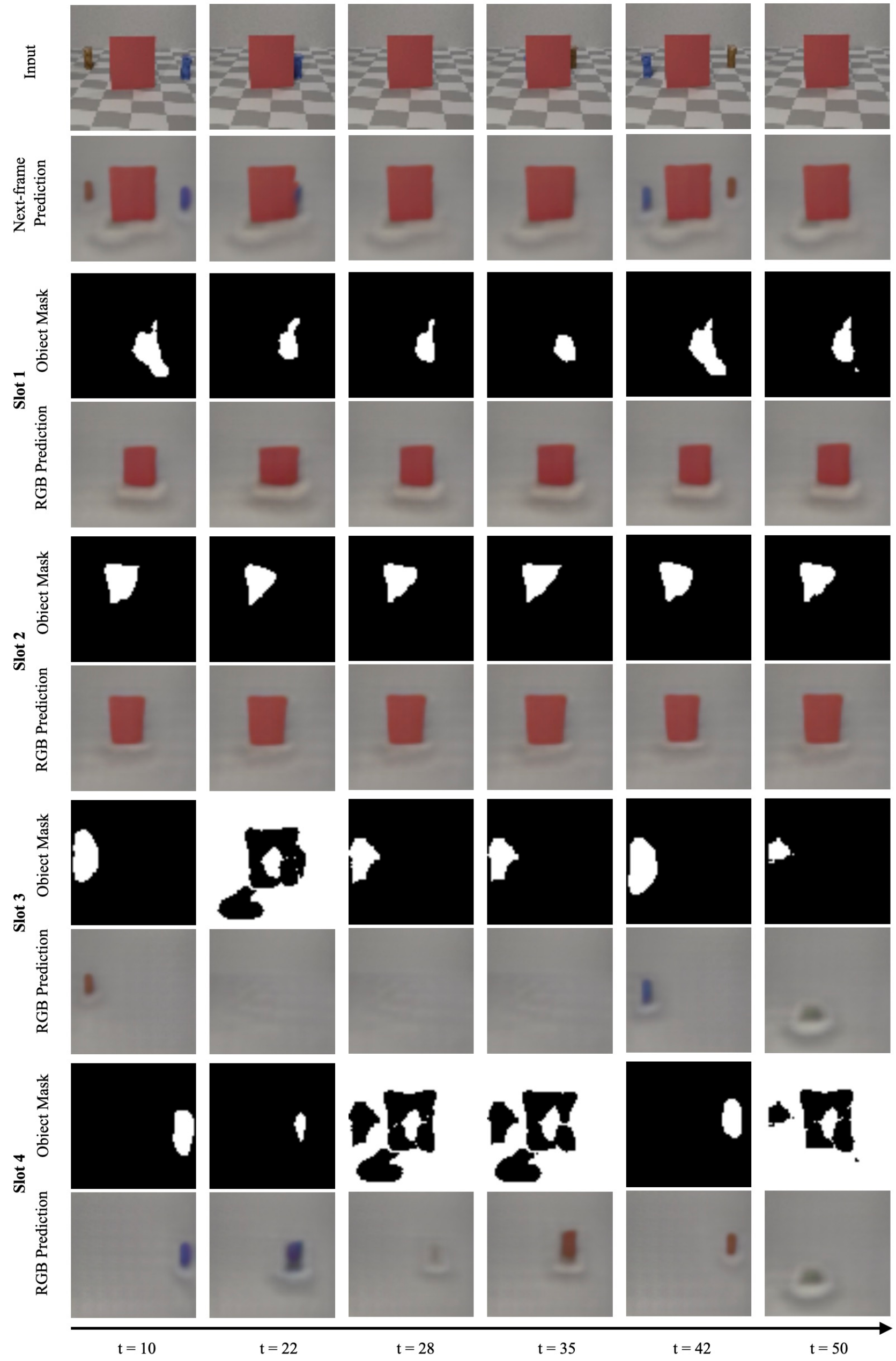}
    \caption{\textbf{The control sequence of the Violation of Expectation experiment and SAVi's perception of it.} (4 out of 7 slots displayed) \textit{Input:} The current frame of the sequence which serves as input. \textit{Next-frame prediction} SAVi's composed RGB prediction for next timestep. \textit{Slot-wise object mask:} Using equation \ref{eq:objectmask} we compute the object masks, depicting full object shapes, the same way as in \locil* \textit{Slot-wise RGB prediction:} SAVi's predicted reconstruction of the object in pixel-space.}
    \label{fig:app:controldemo_savi}
\end{figure*}

\begin{figure*}[h!]
    \centering
    \includegraphics[width=0.9\linewidth]{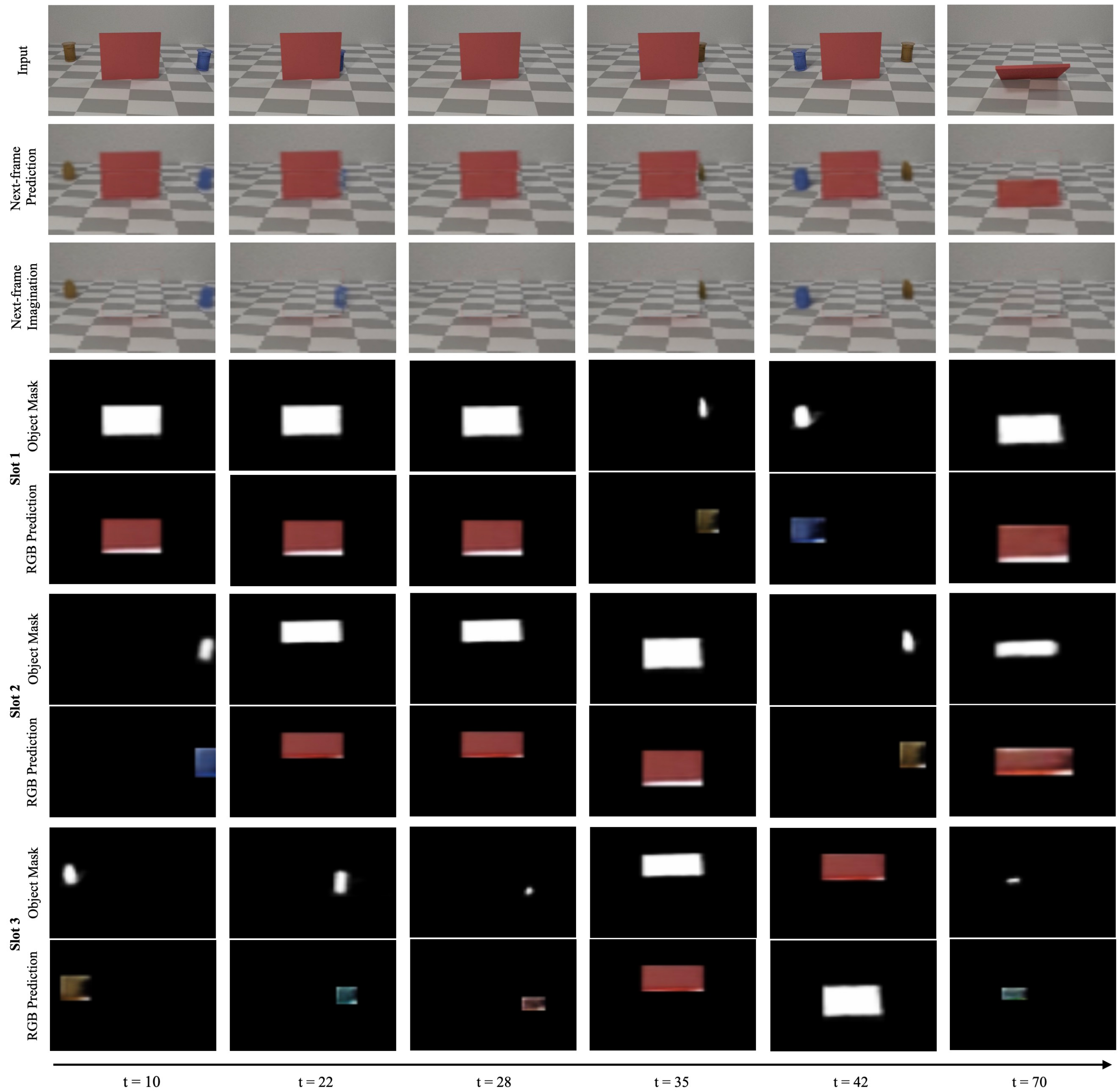}
    \caption{\textbf{The control sequence of the Violation of Expectation experiment and G-SWM's perception of it.}  (3 out of 10 slots displayed) \textit{Input:} The current frame of the sequence which serves as input. \textit{Next-frame prediction} G-SWM's composed RGB prediction for next timestep. \textit{Slot-wise object mask:} Object masks, depicting full object shapes. \textit{Slot-wise RGB prediction:} G-SWM's predicted reconstruction of the object in pixel-space.}
    \label{fig:app:controldemo_gswm}
\end{figure*}

\begin{figure*}[h!]
    \centering
    \includegraphics[width=0.8\linewidth]{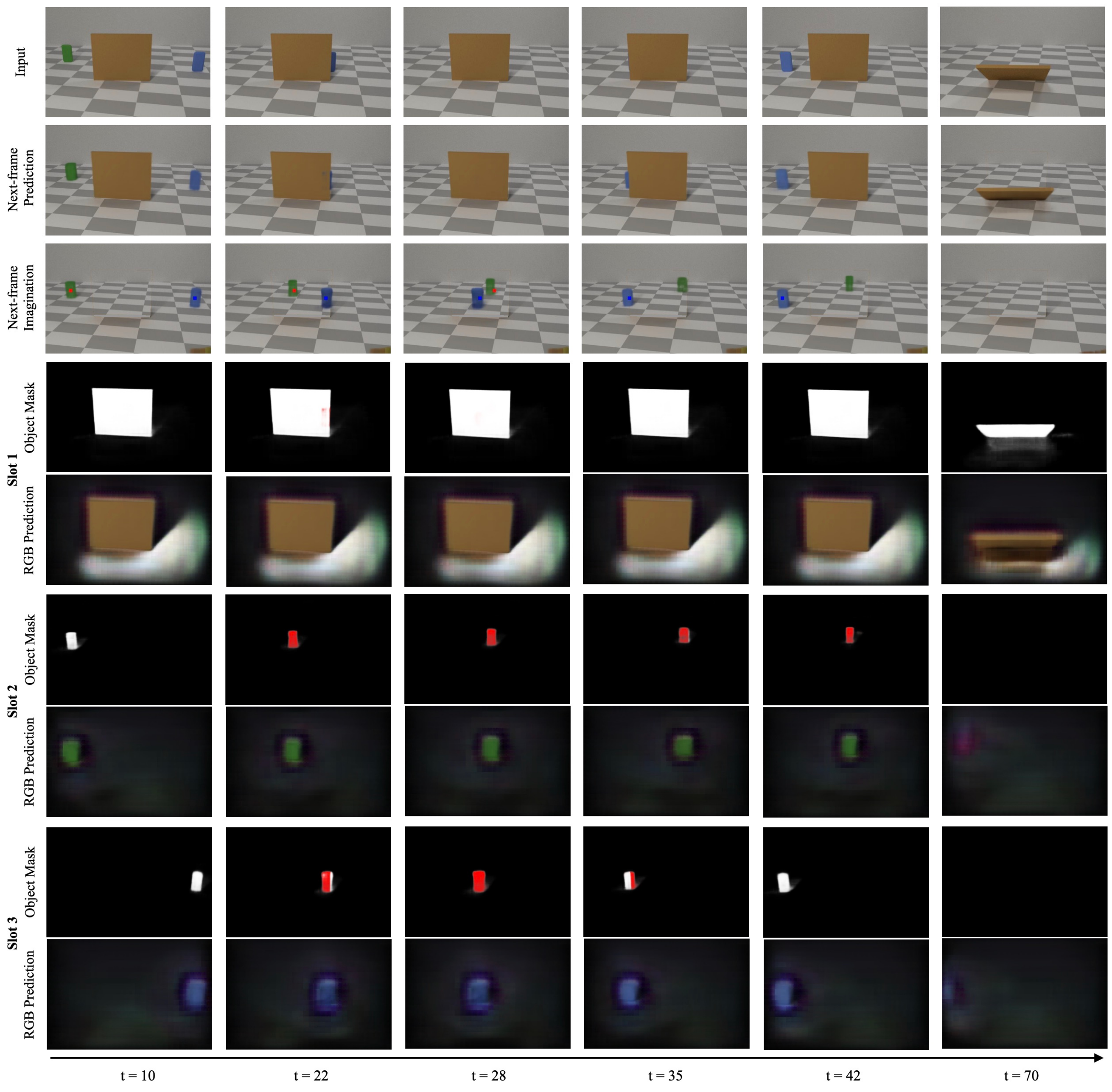}
    \caption{\textbf{The surprise sequence of the Violation of Expectation experiment and \locil*'s perception of it.} \locil* maintains clear object representations throughout the occlusion phase, even when the vanished object does not reappear when initially expected. \textit{Input:} The current frame of the sequence which serves as input. \textit{Next-frame prediction} \locil*'s composed RGB prediction for next timestep. \textit{Next-frame Imagination} \locil*'s composed RGB prediction for next timestep without the occluder screen. The colored dots illustrate the GT positions of the objects. \textit{Slot-wise object mask:} \locil*'s predicted object masks depict full object shapes. Red colored parts correspond to occluded object parts and white colored parts to visible object parts. \textit{Slot-wise RGB prediction:} \locil*'s predicted reconstruction of the object in pixel-space.}
    \label{fig:app:controldemo}
\end{figure*}

\begin{figure*}[h]
    \centering
    \includegraphics[width=0.8\linewidth]{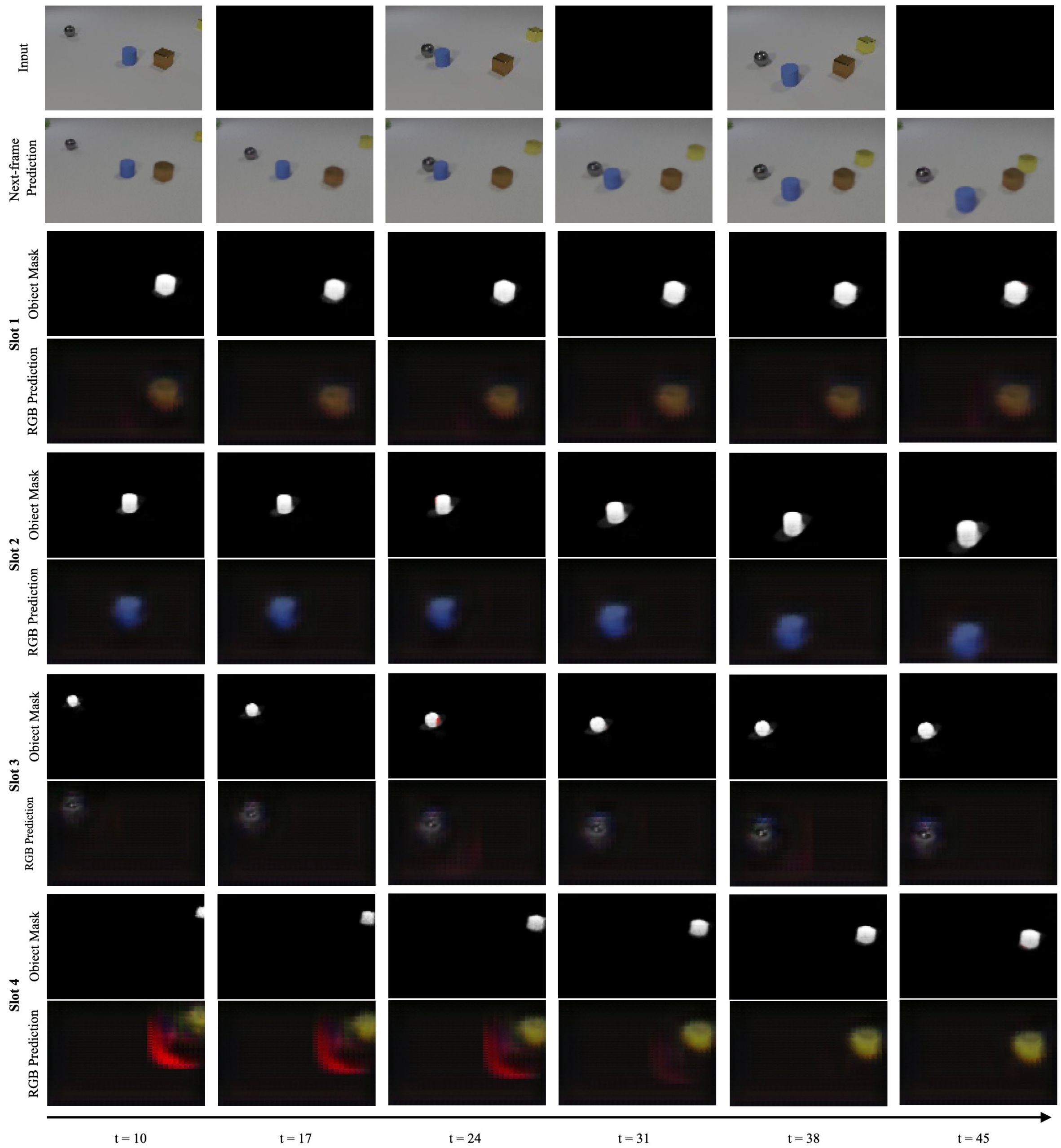}
    \caption{\textbf{The CLEVRER dataset annotated with blackouts and \locil*'s perception of it.} \locil* maintains clear object representations throughout the blackout phases. \textit{Input:} The current frame of the sequence which serves as input. \textit{Next-frame prediction} \locil*'s composed RGB prediction for next timestep. \textit{Slot-wise object mask:} \locil*'s predicted object masks depict full object shapes. Red colored parts correspond to occluded object parts and white colored parts to visible object parts. \textit{Slot-wise RGB prediction:} \locil*'s predicted reconstruction of the object in pixel-space.}
    \label{fig:app:clevrerdemo}
\end{figure*}

\begin{figure*}[h]
    \centering
    \includegraphics[width=0.8\linewidth]{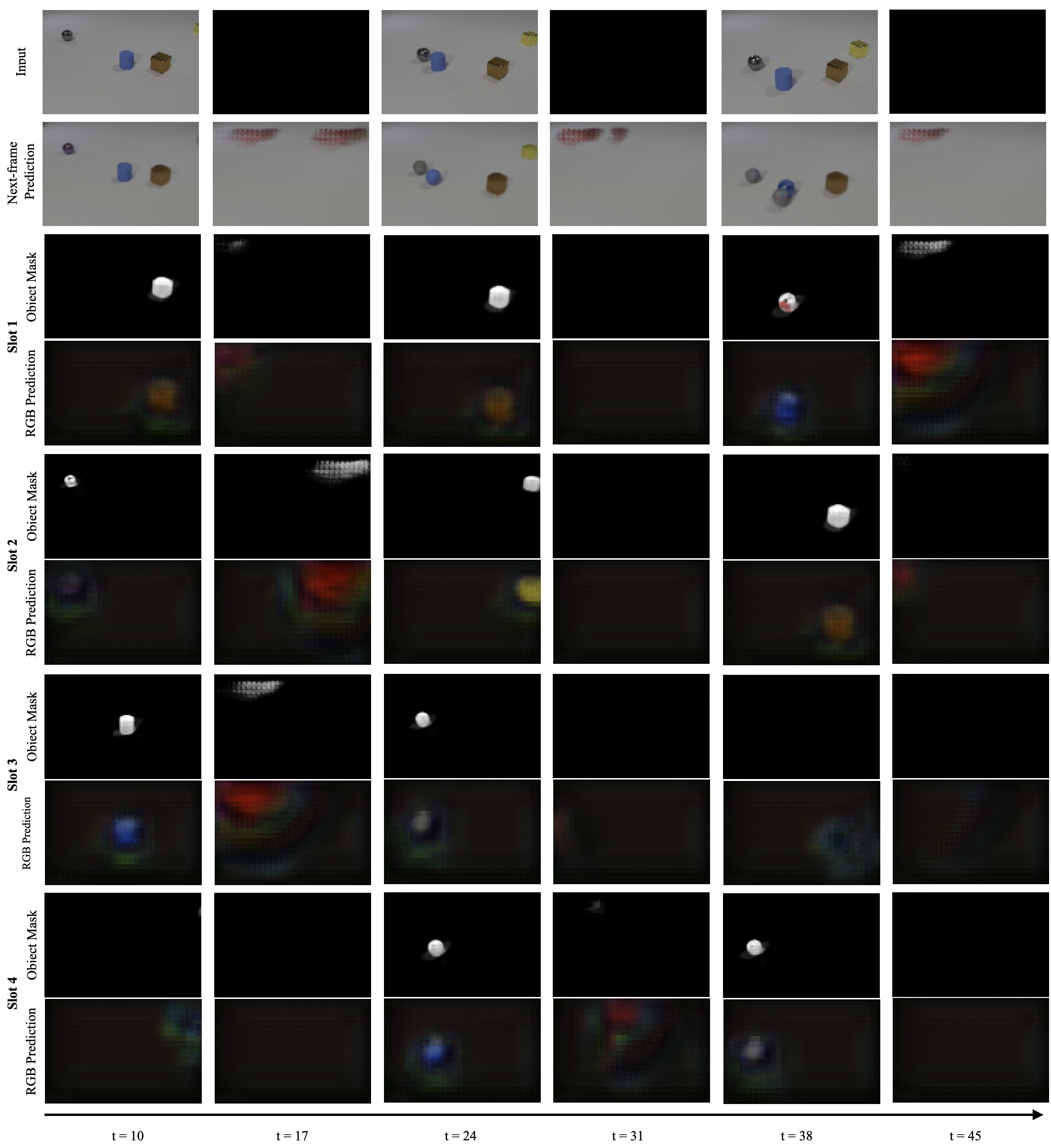}
    \caption{\textbf{The CLEVRER dataset annotated with blackouts and Loci-Unlooped's perception of it.}  \textit{Input:} The current frame of the sequence which serves as input. \textit{Next-frame prediction} Loci-Unlooped's composed RGB prediction for next timestep. \textit{Slot-wise object mask:} Loci-Unlooped's predicted object masks depict full object shapes. Red colored parts correspond to occluded object parts and white colored parts to visible object parts. \textit{Slot-wise RGB prediction:} Loci-Unlooped's's predicted reconstruction of the object in pixel-space.}
    \label{fig:app:clevrerdemo}
\end{figure*}

\begin{figure*}[h]
    \centering
    \includegraphics[width=0.7\linewidth]{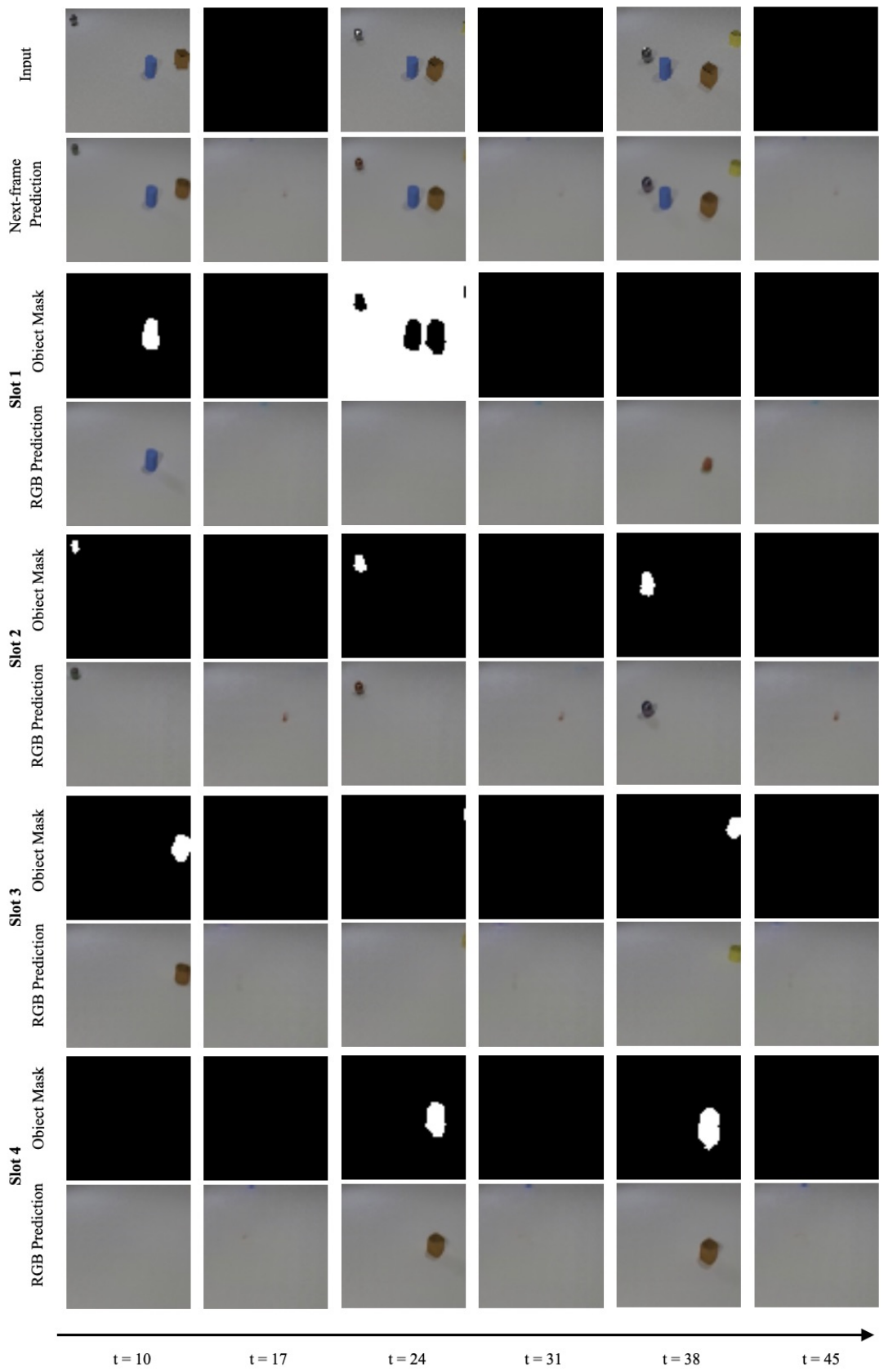}
    \caption{\textbf{The CLEVRER dataset annotated with blackouts and SAVi's perception of it.} (4 out of 7 slots displayed) \textit{Input:} The current frame of the sequence which serves as input. \textit{Next-frame prediction} SAVi's composed RGB prediction for next timestep. \textit{Slot-wise object mask:} Using equation \ref{eq:objectmask} we compute the object masks, depicting full object shapes, the same way as in \locil* \textit{Slot-wise RGB prediction:} SAVi's predicted reconstruction of the object in pixel-space.}
    \label{fig:app:clevrerdemo}
\end{figure*}


\end{document}